\documentclass[10pt,twocolumn,letterpaper]{article}

\usepackage[]{cvpr} 

\usepackage{times}   
 
\usepackage{epsfig}
\usepackage{graphicx}
\usepackage{amsmath}
\usepackage{amssymb}
\usepackage{booktabs}
\usepackage{multirow, xspace}
\usepackage{paralist}
\usepackage{algorithm}
\usepackage{microtype}
\usepackage{xcolor,colortbl}
\usepackage{algorithm}
\usepackage{mathtools}
\usepackage[noend]{algpseudocode}
\usepackage{comment}
\usepackage{dsfont}

\graphicspath{{./}{./figures/}}
\def\Vec#1{{\boldsymbol{#1}}}
\def\Mat#1{{\boldsymbol{#1}}}

\definecolor{aliceblue}{rgb}{0.94, 0.97, 1.0}

\usepackage[normalem]{ulem}
\usepackage{xcolor,colortbl}

\usepackage[pagebackref=true,breaklinks=true,colorlinks,bookmarks=false,citecolor=blue,linkcolor=blue]{hyperref}

\def\cvprPaperID{5542} 
\def\confName{CVPR}
\def\confYear{2022}
\begin{document}
\title{On Generalizing Beyond Domains in Cross-Domain Continual Learning}

\author{Christian Simon$^{1,3,5}$ ~~~~  Masoud Faraki$^2$ ~~~~  Yi-Hsuan Tsai$^6$  ~~~~  Xiang Yu$^2$ \\
Samuel Schulter$^2$ ~~~~ Yumin Suh$^2$ ~~~~ Mehrtash Harandi$^{3,5}$  ~~~~ Manmohan Chandraker$^{2,4}$ \\ 
\\
$^1$The Australian National University  ~~ $^2$NEC Labs America  ~~ $^3$Monash University\\
 $^4$University of California, San Diego  ~~~ $^5$Data61 ~~~~ $^6$Phiar Technologies\\
{\tt\small sen.christiansimon@gmail.com, mfaraki@nec-labs.com, wasidennis@gmail.com}\\ {\tt\small \{xiangyu, samuel, yumin\}@nec-labs.com, mehrtash.harandi@monash.edu, manu@nec-labs.com}
}

\maketitle

\begin{abstract}
Humans have the ability to accumulate knowledge of new tasks in varying conditions, but deep neural networks often suffer from catastrophic forgetting of previously learned knowledge after learning a new task. Many recent methods focus on preventing catastrophic forgetting under the assumption of train and test data following similar distributions. In this work, we consider a more realistic scenario of continual learning under domain shifts where the model must generalize its inference to an unseen domain. To this end, we encourage learning semantically meaningful features by equipping the classifier with class similarity metrics as learning parameters which are obtained through Mahalanobis similarity computations. Learning of the backbone representation along with these extra parameters is done seamlessly in an end-to-end manner. In addition, we propose an approach based on the exponential moving average of the parameters for better knowledge distillation. We demonstrate that, to a great extent, existing continual learning algorithms fail to handle the forgetting issue under multiple distributions, while our proposed approach learns new tasks under domain shift with accuracy boosts up to 10\% on  challenging datasets such as DomainNet and OfficeHome.  
\end{abstract}


\section{Introduction}

\begin{figure}[t]
    \centering
    \includegraphics[width=0.48\textwidth]{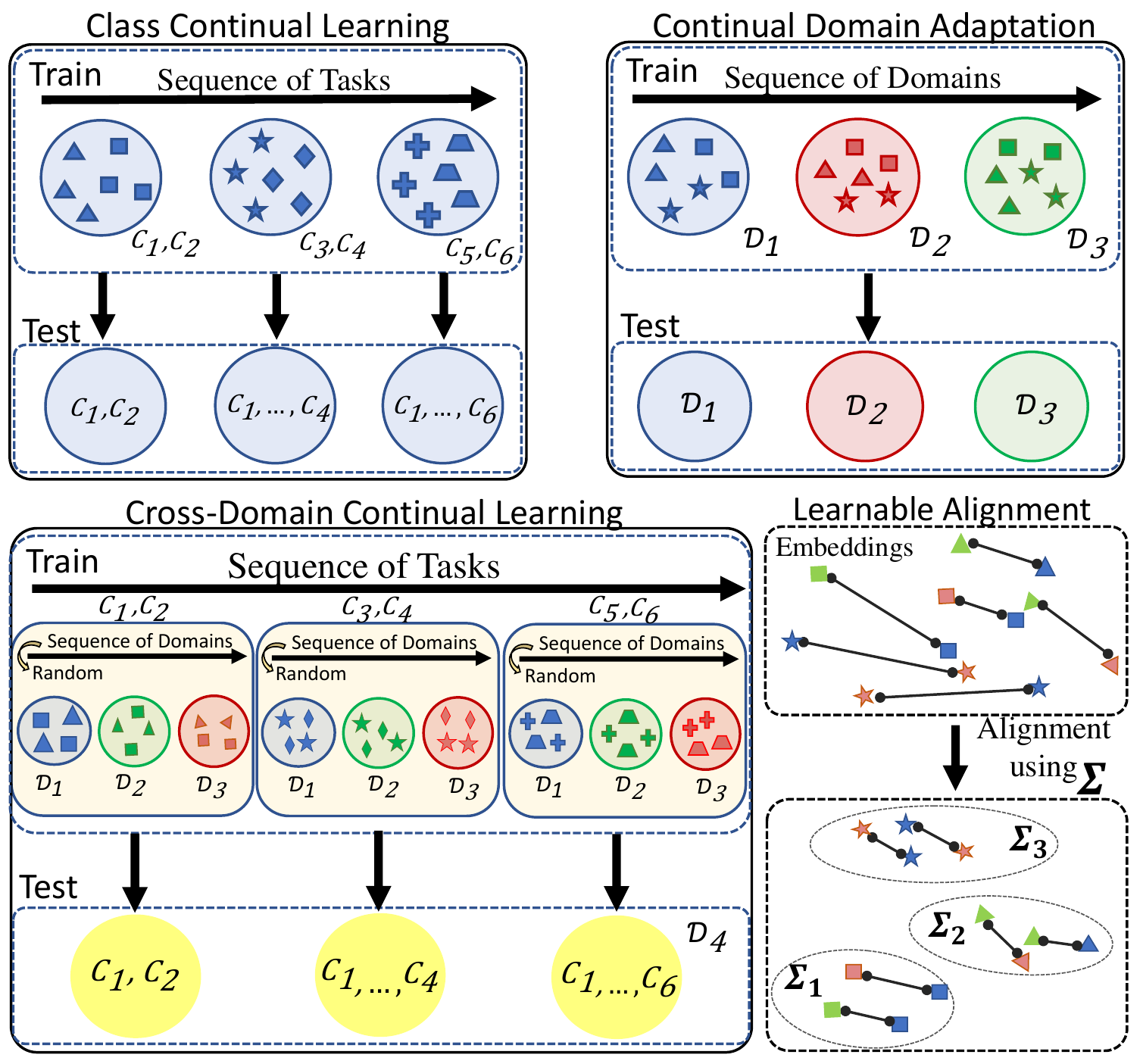}
    \caption{ 
      Top: Existing settings on 1) continually learning new visual categories from single domain (left), and 2) continually learning from new domains with evaluation on the same domains (right). Bottom: Our setting which has a sequence of visual categories coming from various domains, with evaluation on an unseen domain. The proposed approach utilizes a continual domain alignment strategy dubbed Mahalanobis Similarity Learning (MSL). Colors indicate domains, and shapes indicate categories.
    }
	\label{fig:teaser}
	\vspace{-0.2cm}
\end{figure}

Humans possess the extraordinary capability of acquiring new knowledge in dynamically changing environments, while preserving knowledge learned in the past. The obtained knowledge can be further generalized to unseen situations without the need of re-educating. On the other hand, there has been a surge of efforts to devise machine learning based algorithms to build more intelligent models and mitigate the aforementioned challenges from two perspectives, namely continual learning~\cite{masana2020classincsurvey,belouadah2019il2m,castro2018end,van2019three} and domain generalization~\cite{gulrajani2021domainbed,li2017pacs,Li2018MLDG,Ghifary2017ScatterCA}. This is particularly more important when deployed in the real world under a life-long learning setup~\cite{rebuffi2017icarl,kirkpatrick2017ewc,lopez2017gem}. For instance, consider warehouse robots that might perceive new inventory or unseen room layouts that they require to adapt to function properly. The observations are captured at different time frames (\eg, day or night) and different locations (\eg, aisles) such that the observed domains come with an unpredictable sequence.  In these situations, the key to success is to have certain embedded adaptability in the robots to handle the challenges without costly re-training or entirely replacing them.

To put the discussion into perspective, on one hand, continual learning based methods mainly try to deal with \textit{catastrophic forgetting}, which refers to the performance degradation of previously acquired knowledge when new concepts are learned. On the other side, domain generalization is to find a good feature representation that goes well beyond the training distributions, while at the same time being discriminative for the task at hand. While effective, there has been comparatively little efforts in research to provide answers for the two aforementioned challenges simultaneously. One effort is the work of Volpi~\etal~\cite{volpi2021continual} which proposes continual domain adaptation, \ie, where different domains arrive in a continual fashion (top-right in Fig.~\ref{fig:teaser}).  Other similar effort includes the work of Kundu~\etal~\cite{kundu2020class}, which suggests class-continual learning with source-target domain adaptation as in the open-set setting. However, in both works the main aspect of \emph{generalization} beyond seen domains is largely missing, limiting applicability of them in real-world scenarios. Moreover, the notion of incrementally adding training tasks is constrained to source and target domains only (\ie, two tasks).

In this work, we propose an approach for cross-domain continual learning, which also has the capability of generalization to unseen domains. 
Our setup considers a sequence of tasks (\ie different visual categories), where each task's data is originated from various domains (Fig.~\ref{fig:teaser} bottom-left). 
Note that, our setup does not have any prior assumptions about the domains (\eg, availability of domain identifiers or specific orderings) associated to the given training samples in each task.  
This is a realistic scenario where the model is deemed to be agnostic about the origin of training samples, \eg, when preserving privacy is important.  
We deem the domain alignment be done in a discriminative manner by equipping our classifier with class-specific Mahalanobis similarity metrics, as shown in Fig.~\ref{fig:teaser} (bottom-right). Here, the classifier network also takes into account the underlying distribution of the class samples when generating the predictions. This is to encourage learning semantically meaningful features across training domains. We then learn the backbone representation parameters along with these extra parameters in an end-to-end fashion. In addition, we propose an approach based on the exponential moving average of the parameters for better knowledge distillation, preventing excessive divergence from the previously learned parameters.

To evaluate our method, we define highly dynamic environments with data coming from various domains and expanding visual categories.  We perform extensive experiments on four different datasets -- DomainNet~\cite{peng2019moment}, OfficeHome~\cite{venkateswara2017officehome}, PACS~\cite{li2017pacs}, and NICO~\cite{he2020nico}.  The results show that our method consistently leads to an improvement of up to 10\% compared to baselines~\cite{zhou2021mixstyle,li2017lwf,simon2021geodl,hou2019lucir,zhang2020arm} on 10-task and 5-task protocols. Furthermore, our proposed method also prevents \emph{catastrophic forgetting}, achieving the lowest backward transfer rate~\cite{lopez2017gem} on average, \eg, $\sim$10\% and $\sim$8\% on DomainNet and OfficeHome, respectively.

To summarize, our contributions include
\begin{enumerate}
    \item  We provide a unified testbed for cross-domain continual learning with comparison to continual learning methods and techniques for domain generalization. 
    \item  We propose a projection technique in an end-to-end scheme for domain generalization. In particular, we make use of learnable Mahalanobis similarity metrics for robust classifiers against unseen domains.
    \item  We devise an exponential moving average framework for knowledge distillation. The proposed module is integrated with our learnable projection technique to alleviate the degrading impact of {catastrophic forgetting} and distributional shifts by adaption to a history of the old parameters.
\end{enumerate}

\section{Related Work}
\noindent{\textbf{Continual Learning.}} To tackle the forgetting problem in continual learning, neural networks  must maintain the performance of past visual categories. Knowledge Distillation~\cite{Hinton2015Distillation, buzzega2020dark} (KD) between an old model and a current model is an effective approach to prevent {catastrophic forgetting}.
The standard baseline exploits KD as proposed by Li and Hoiem~\cite{li2017lwf}, for which the predictions between old and  current models are preserved. Hou \etal~\cite{hou2019lucir} propose a KD method in feature space, in order to maintain the features from old and current models. In the same line of research, Simon \etal~\cite{simon2021geodl} introduce a smooth property to learn from one task to another task such that the geometry aspect is taken into account.  %
Another stream of continual learning methods consider memory selection and generation for memory replay. A classical approach is known as herding~\cite{welling2009herding} by picking the nearest neighbors from the mean of exemplars in each class. Another approach in this category is gradient episodic memory~\cite{lopez2017gem,chaudry2019agem} that uses old training data to impose optimization constraints when learning new tasks. Liu \etal~\cite{liu2020mnemonics} use bi-level optimization for synthesizing the memory, and a more optimal memory replay is expected compared to storing exemplars from training data.
Despite the wide use of generating and selecting exemplars for continual learning, it does not guarantee robustness under change of train and test distributions.

\vspace{0.02cm}
\noindent\textbf{Domain Generalization}.
Domain generalization techniques aim to generalize beyond training domains, which is a different goal compared to domain adaptation that reduces the distributional shifts between source and target domains. The problem of domain generalization also differs from few-shot or unsupervised domain adaptation where in these problems the test data is accessible during training~\cite{wilson2018daunsup,gannin2015unsupbackprop}. A standard approach is to expose a model with various domains in training as recommended in~\cite{Vapnik1998} under empirical risk minimization. This straightforward idea from supervised learning is effective for domain generalization as also shown in~\cite{gulrajani2021domainbed}. An extension is to adapt the risk minimization loss to a context network with a meta learning strategy as proposed in~\cite{zhang2020arm}. To improve generalization, Zhou \etal~\cite{zhou2021mixstyle} propose a smooth style transfer applied to the feature statistics. Though these techniques are effective to generalize to unseen distributions, their ability to deal with a stream of data containing multiple tasks is still questionable.

\vspace{0.02cm}
\noindent \textbf{Learning Embeddings}. To compute the distance between a pair of points, a projection matrix (\eg, covariance, Positive Semi-Definite matrices) plays a crucial role in image recognition. Bardes \etal~\cite{bardes2021vicreg} apply covariances for correlation and decorrelation among samples to avoid \textit{collapse} (\ie non-informative feature vectors). Faraki \etal~\cite{faraki2021cross} propose a cross-domain triplet loss using covariances for domain alignment. The projection matrices are also known to be effective to compute similarity between two entities as proposed in~\cite{zhao2018domaininvariant,simon2020adaptive}. In comparison, our proposed approach employs discriminative projection matrices for the learned features in a form of Mahalanobis metrics and bias terms to generate robust predictors.

\section{Proposed Method}
\label{sec3:Proposed}
In this section, we present our approach to learning tasks sequentially with: \textbf{1}) constraints on the storage of the previously observed learning samples, and \textbf{2}) severe distribution shifts within the learning tasks, without suffering from the so-called issue of \textit{catastrophic forgetting}. Our learning scheme identifies the feature and similarity metric learning jointly. In particular, we learn class-specific similarity metrics defined in the latent space to increase the discriminatory power of features in the space. This is done seamlessly along with learning the features themselves. 

Below, we first review some basic concepts used in our framework. Our method tackles the domain generalization and {catastrophic forgetting} by incorporating two components: \textbf{1}) domain generalization strategy by learning Mahalanobis metrics and \textbf{2}) preserving past knowledge based on knowledge distillation using exponential moving average of parameters, which are discussed afterwards. 

\subsection{Notation and Preliminaries} 
\label{sec:notation}
Throughout the paper, we denote vectors and matrices in bold lower-case (\eg, $\Vec{x}$) and bold upper-case letters (\eg $\Mat{X}$), respectively. On $\Vec{x}$, $[\Vec{x}]_{i}$ denotes the element at position ${i}$ while $\|\Vec{x}\|^2_2=\Vec{x}^\top\Vec{x}$ shows its squared $l_2$ norm. We denote a set by $\mathcal{S}$.

Formally, in continual learning, a model is trained in several steps called \textit{tasks}. Each task $T_i, 1 \leq i \leq q$, consists of samples of a set of novel classes $\mathcal{Y}_i^N$ as well as samples of a set of old classes $\mathcal{Y}_i^O$. The aim is to train a model to classify all seen classes, \ie, $\mathcal{Y}_i^O \cup \mathcal{Y}_i^N$. The allowed number of training samples for $\mathcal{Y}_i^O$ is severely constrained (called \textit{rehearsal memory} $\mathcal{M}$).

\begin{figure}[t]
	\centering
	\includegraphics[width=0.479\textwidth]{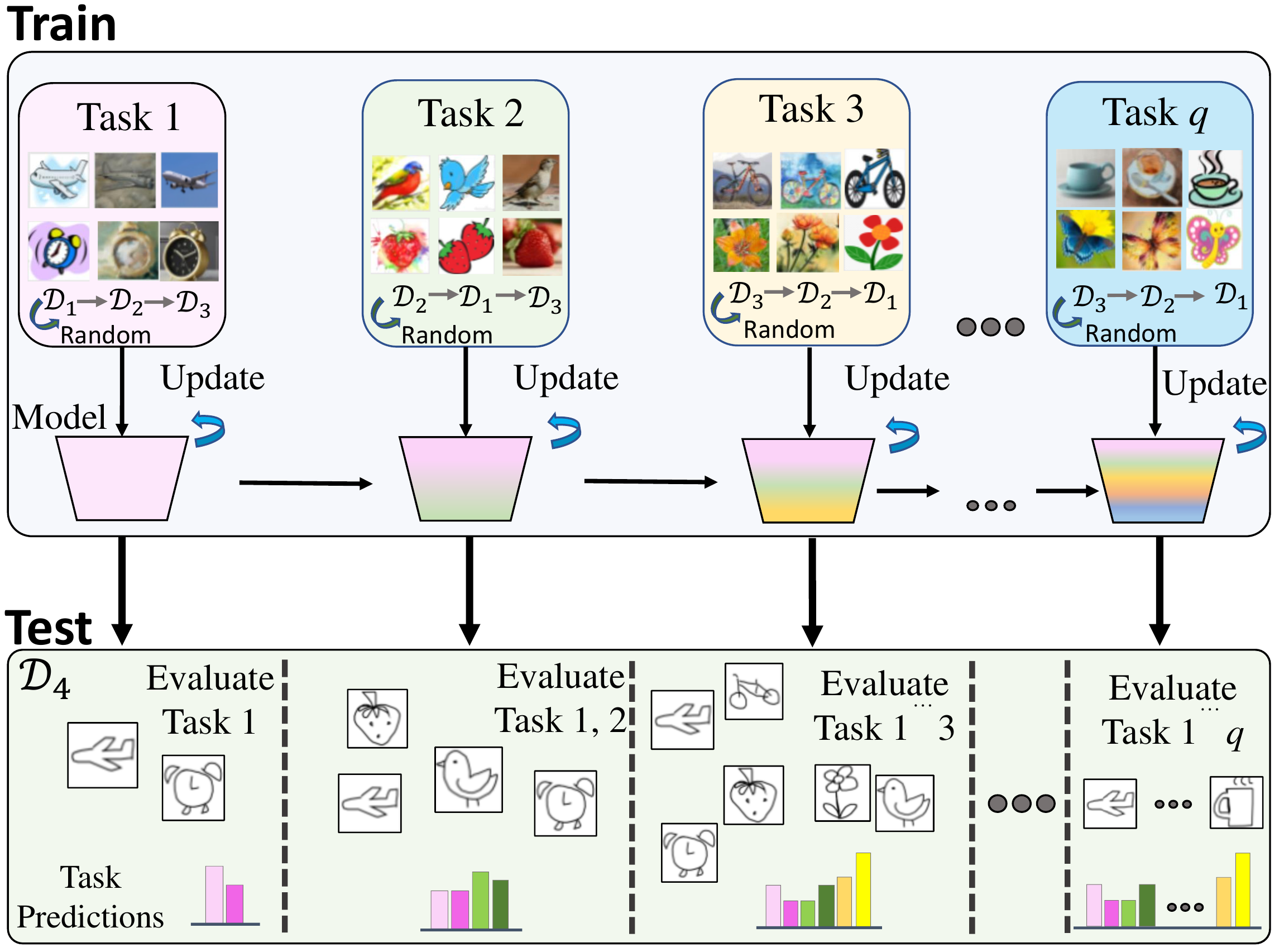}
	\caption{The overall setting in cross-domain continual learning. The training problem is divided into several tasks, where each new task has a subset of novel object categories coming from various training domains. While the training data from old tasks is discarded at each time, the model has to learn sequentially from the incoming tasks to evaluate on inputs from an unseen domain with a different distribution.
	}
	\label{fig:setting}
\end{figure}

In our cross-domain continual learning setup, we tackle the recognition scenario where during training we observe $m$ source domains, \ie,  $\mathcal{D}_1, \dots, \mathcal{D}_{m}$, each with different distributions. The learning sequence is defined as learning through a stream of tasks $T_1, \dots, T_q$, where the data from each task is composed of a sequence of $m$ source domains. 
Note that, in our setup, we do not require the information about the domains (\eg, domain identifier) from which the samples in each task are given. When feeding the training data in each episode, we are interested in averaging the performance measures when the data domains are in random orders and the process is repeated for a number of times (\eg, 5). %
Like the standard continual learning setup, knowledge from a new set of classes  is learned from each novel task. 
At the test time, we follow the domain generalization evaluation protocol in which the trained model has to predict $y \in {\bigcup}_{i=1}^{q} \mathcal{Y}_i$,
values of inputs from an unseen/target domain $\mathcal{D}_{m+1}$. We note that $\mathcal{D}_{m+1}$ has samples from an unknown distribution. Our setting is presented conceptually in Fig.~\ref{fig:setting}.

Like a standard continual learning method, we also apply experience replay by storing exemplars in the memory $\mathcal{M}$. To some extent, this would help preventing the forgetting issue. The exemplars stored in the memory are constructed from each class and each domain. %
{We store randomly selected exemplars in the memory and ensure that every run uses this same set of exemplars.} %
In the following, for simplicity, we drop the task indicator $i$ and assume the size of label space is $C$. 

\begin{figure*}[t]
	\centering
	\includegraphics[width=1.01\textwidth]{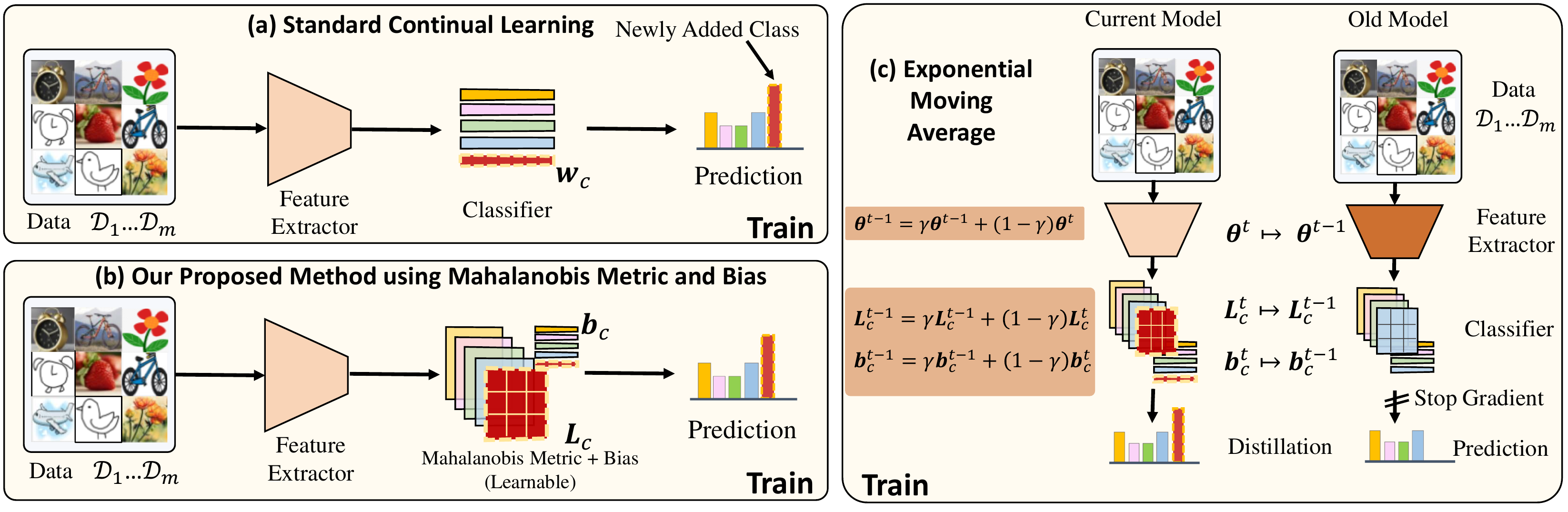}
	\caption{The pipeline of our approach. (a) For comparison, we show a standard continual learning approach with expanding parameters when a new class is presented. 
	(b) Our approach also expands the classifier with Mahalanobis metrics and biases as learnable parameters to learn semantically meaningful features across training domains. (c) Extension of our proposed domain generalization method with knowledge distillation to allow smooth updates when learning new tasks. }%
	\label{fig:pipeline}
\end{figure*}

\subsection{Domain Generalization by Learning Similarity Metrics}
In this part, we present our approach to learn class similarity metrics in a cross-domain continual learning setting, with the focus of generalization to unseen domains. To this end, we encourage learning semantically meaningful features by equipping the classifier with class similarity metrics which are obtained through Mahalanobis similarity computations. Here, we deem the domain alignment be done in a discriminative manner. In doing so, our idea is consistent with recent works that utilize a notion of feature semantics in their domain alignment inference to avoid undesirable effects of aligning semantically different samples from different domains.
To name a few, the Contrastive Adaptation Network (CAN) for unsupervised domain adaptation~\cite{kang2019contrastive}, the Covariance Metric Networks (CovaMNet) for few-shot learning~\cite{li2019distribution}, the Model-Agnostic learning of Semantic Features (MASF) for standard domain generalization~\cite{dou2019domain} and the Cross-Domain Triplet (CDT) loss for face recognition from unseen domains~\cite{faraki2021cross}. %

We begin by introducing the overall network architecture. Our architecture closely follows a typical image recognition design used in continual learning setting. Let $f_{\theta} : \mathcal{X} \rightarrow \mathcal{H}$ represents a backbone CNN parametrized by $\theta$ which provides a mapping from the image input space $\mathcal{X}$ to a latent space. Furthermore, let $f_{\phi} : \mathcal{H}  \rightarrow \mathcal{Y}$ be a classifier network parametrized by $\phi$ that maps the outputs of $f_{\theta}$ to class label values. More specifically, forwarding an image $x$ through $f_\theta(\cdot)$ outputs a tensor that, after being flattened (\ie, $f_\theta(x) \in \mathbb{R}^n$ )
acts as input to the classifier network $f_\phi(\cdot)$. In a typical pipeline, the goal is to train a model on each task $T_i, 1 \leq i \leq q$, while expanding the output size of the classifier to match the number of classes. Note that the sequential learning protocol in our setting does not have strong priors and assumptions \eg, domain identities and overlapping classes.  

In most continual learning methods~\cite{hou2019lucir, simon2021geodl, li2017lwf}, the classifier network $f_\phi$ is often implemented by a Fully-Connected (FC) layer with weight $\Mat{W} = [\Vec{w}_1, \dots ,\Vec{w}_C]^\top \in \mathbb{R}^{C \times n}$
with $\Vec{w}_i \in \mathbb{R}^n$. 
When learning a new task, $\Mat{W}$ is expanded to cover $k$ new task categories by accommodating $k$ new rows, \ie, $\Mat{W} = [\Vec{w}_1, \dots ,\Vec{w}_{C},\Vec{w}_{C+1}, \dots, \Vec{w}_{C + k}]^\top$. A similarity score between a class weight $\Vec{w}_c$ and a feature $f_\theta(x) = \Vec{h} \in \mathbb{R}^{n}$ associated with an image $x$ is then defined by projection  as $\langle \Vec{w}_c, \Vec{h} \rangle = \Vec{w}_c^\top \Vec{h}$ to be optimized by a loss function (see Fig.~\ref{fig:pipeline} (a)). Despite its wide use, we argue that this approach is not robust to distributional shifts as it is not explicitly designed to align samples that are seen in previous classes but from different distributions. 

Here, we deem the classifier network also take into account the underlying distribution of the class samples when generating the class predictions. To this end, we equip the classifier network with Positive Semi-Definite (PSD) Mahalanobis similarity metrics $\Mat{\Sigma}_c$ as learnable parameters, to encourage learning semantically meaningful features across different domains. Furthermore, category features are allowed to shift by learning a bias vector $\Vec{b}_c$. We store these parameters in the network and expand to match the number of new classes when learning a new task. Therefore, after learning a new task, the prediction layer in our framework consists of extra learnable parameters $\phi = \{\Mat{\Sigma}_1, \Vec{b}_1, \dots, \Mat{\Sigma}_{C},\Vec{b}_{C}\}$. We then learn the backbone representation parameters $\theta$ along with $\phi$ in an end-to-end manner. 
 
Utilizing $\phi$, the proposed similarity score with respect to class $c$ for an image $x$ passing through the network can be obtained by 
\begin{equation}
    \mathrm{sim}_c({x}; \theta, \phi) =  (f_\theta({x}) - \Vec{b}_c)^\top \Mat{\Sigma_c}~ (f_\theta({x}) - \Vec{b}_c).
    \label{eqn:mahalanobis}
\end{equation}

\paragraph{Intuition.}
The motivation behind Mahalanobis similarity learning is to determine $\Mat{\Sigma}_c$ such that by learning to expand or shrink axes of $f_\theta({x}) \in \mathbb{R}^{n}$, certain useful properties are achieved when generating \eqref{eqn:mahalanobis}. To better understand the behavior of our learning algorithm, let $\Vec{r_c}=(f_\theta({x}) - \Vec{b}_c)$ and the eigendecomposition of $\Mat{\Sigma}_c$ be $\Mat{\Sigma}_c = \Mat{V}_c \Mat{\Lambda}_c \Mat{V}_c^\top$. Then,
\begin{align} \notag
\Vec{r_c}^\top \Mat{\Sigma_c} \Vec{r_c} &= \big( \Mat{\Lambda_c}^\frac{1}{2} \Mat{V_c}^\top \Vec{r_c} \big)^\top  \big( \Mat{\Lambda_c}^\frac{1}{2} \Mat{V_c}^\top \Vec{r_c} \big) \notag \\ 
&=  \big\|  \Mat{\Lambda_c}^\frac{1}{2} \Mat{V_c}^\top \Vec{r_c} \big\|^2_2,
\label{eqn:intuition}
\end{align}
which associates $\Vec{r_c}$ with the eigenvectors of $\Mat{\Sigma_c}$ weighted by the eigenvalues. When $\Vec{r_c}$ is in the direction of leading eigenvectors of $\Mat{\Sigma_c}$, it obtains its maximum value. Then, optimizing this term over the associated class samples leads to a more discriminative alignment of the data sources.

\paragraph{A computationally more efficient alternative.} Taking advantage of the structure of $\Mat{\Sigma}$, we can further decompose it to have a more efficient version. The similarity metric matrix can be decomposed to $\Mat{\Sigma}_c  = \Mat{L}_c^\top \Mat{L}_c$, where $\Mat{L} \in \mathbb{R}^{r \times n}$ with $r \ll n$. This will ensure $\Mat{\Sigma}$ remains PSD and yields valid similarity scores~\cite{faraki2018comprehensive,faraki2017large}. Furthermore, it can substantially reduce storage needs and increase the scalability of our method when a large-scale application is deemed. In practice, this lets us conveniently implement $\Mat{\Sigma}$ by a FC layer into any neural network.
Using the decomposition, \eqref{eqn:mahalanobis}  boils down to
\begin{align}
    \mathrm{sim}_c({x}; \theta, \phi) = \big\|\Mat{L}_c \big(f_\theta ({x}) - \Vec{b}_c\big) \big\|^2_2.
    \label{eq:cov_dist}
\end{align}
\paragraph{Overall training pipeline.} Finally, the updated classifier parameters become $\phi = \{\Mat{L}_1, \Vec{b}_1, \dots, \Mat{L}_{C},\Vec{b}_{C}\}$. Later, in experiments, we will study the effect of different values of $r$ in our framework. We store some examples from seen tasks and various domains. During training, the samples $x$ in mini-batches come from the samples in the current task and the memory. Thus, our objective becomes minimizing the loss function across domains and samples. Our parameters $\phi$ that represents each class are updated during training in conjunction with the feature extractor parameters $\theta$. We train our model using the cross-entropy loss, which is widely used for Empirical Risk Minimization (ERM)~\cite{Vapnik1998}
\begin{align}
   \mathcal{L}_\mathrm{CE} = - \!\!\!\!\!\! \sum_{{x} \in {\mathcal{X} \cup \mathcal{M}}}\!\!\!\!\! \delta_{y=c}  \log \frac{\mathrm{exp}\big(\mathrm{sim_c}({x}; \theta, \Mat{L}_c, \Vec{b}_c)\big)}{\sum_{c^\prime}\mathrm{exp}\big(\mathrm{sim}_{c^\prime}({x}; \theta, \Mat{L}_{c^\prime}, \Vec{b}_{c^\prime})\big)},
   \label{eq:ce_loss}
\end{align}
where $\delta$ is an indicator function corresponding with the label $y$. 

As mentioned earlier, we have a memory of exemplars from various domains. Thus, the learnable parameters can be updated towards a more generalized classifier as an attempt to improve classification on unseen domains. We conceptually show our proposed method for Mahalanobis Similarity Learning (MSL) in Fig.~\ref{fig:pipeline} (b).%

\subsection{Knowledge Distillation with Exponential Moving Average}
In this section, we develop an effective Knowledge Distillation (KD) strategy to take advantage of previously learned knowledge without requiring old tasks' images and labels. While many other methods focus on applying KD~\cite{french2019selfensemble} using only the old and current models~\cite{li2017lwf,buzzega2020dark,simon2021geodl,hou2019lucir}, we utilize a history of previous knowledge to limit the divergence from the old model. Let ${\Psi}^t=\{{\theta}^t, {\phi}^t\}$ and ${\Psi}^{t-1}=\{{\theta}^{t-1}, {\phi}^{t-1}\}$ be all learnable parameters in our framework at the current and old tasks, respectively. %
Then, given a temperature $\tau$, we propose the following KD on the predictions of the current and old models   
\begin{align}
\mathcal{L}_{\mathrm{Dis}}({\Psi}^t,{\Psi}^{t-1}; {x}) = - \sum_{c=1}^{C} {p}^{t-1}_c({x}) \log p^t_c({x}),
\label{eq:distillation_loss}
\end{align}
with 
\begin{equation}
\resizebox{1\hsize}{!}{
     ${p}^{t-1}_c({x}) = \frac{\text{exp}\big({\mathrm{sim}_c({x};\Psi^{t-1})/\tau}\big)}{\sum_{c^\prime=1}^{C} \text{exp}\big({\mathrm{sim}_{c^\prime}({x};\Psi^{t-1})/\tau}\big)}, \; 
     {p}^{t}_c({x}) = \frac{\text{exp}\big({\mathrm{sim}_c({x};\Psi^{t})/\tau}\big)}{\sum_{c^\prime=1}^{C} \text{exp}\big({\mathrm{sim}_{c^\prime}({x};\Psi^{t})/\tau}\big)}$,
     } \nonumber
\end{equation}
where the similarity score, $\mathrm{sim(\cdot)}$, is obtained by \eqref{eq:cov_dist}. 

As also  empirically observed in~\cite{french2019selfensemble}, temporal ensembling methods applied to the old model stabilizes the training. Here, the idea is that the outputs of the current model must not significantly deviate form the old ones.
To this end, we employ a smooth parameter update strategy using Exponential Moving Average (EMA) updates. Connecting to KD, the idea is to smoothly guide the learning of the parameters of the current model while taking into account the predictions from the old ones. Therefore, we define the EMA update in our framework as   
\begin{align}
\begin{split}
    {\theta}^{t-1} &= \gamma {\theta}^{t-1} + (1-\gamma) {\theta}^t,\\
    \Vec{b}_c^{t-1} &= \gamma \Vec{b}_c^{t-1} + (1-\gamma) \Vec{b}^t_c,\\
    \Vec{L}_c^{t-1} &= \gamma \Vec{L}_c^{t-1} + (1-\gamma) \Vec{L}^t_c,
\end{split}    
\label{eq:mov_avg}
\end{align}
where $\gamma$ is a positive smoothing coefficient hyperparameter. 

Furthermore, we apply a stop-gradient operator to the old model. Once the training is done, the old model is discarded.
The process is depicted in Fig.~\ref{fig:pipeline} (c). Our overall loss then becomes $\mathcal{L}_\mathrm{CE} + \lambda \mathcal{L}_\mathrm{Dis}$, with $\lambda$ showing the weight of KD loss. We dub this method MSL + Mov in our experiments. 
%

\section{Experiments}
\label{sec4:Expr}

In this section, we compare and contrast our method against  existing methods in both (class) Continual Learning (CL) and Domain Generalization (DG). We start with introducing our competitor methods and experimental details. 
\paragraph{Baselines.} To evaluate our proposed method, we compare with the competitor methods in CL, namely LwF~\cite{li2017lwf}, LUCIR~\cite{hou2019lucir}, and GeoDL~\cite{simon2021geodl}.
Concisely, LwF~\cite{li2017lwf} applies knowledge distillation on the predictions, while LUCIR~\cite{hou2019lucir} employs preservation of features from old and current models, and GeoDL~\cite{simon2021geodl} extends the preservation in the feature space with smooth transitions of two subspaces from the old and the current models.  These models are a combination of widely used and very recent methods. Furthermore, we include a baseline Empirical Risk Minimization (ERM)~\cite{Vapnik1998} as well as recent DG methods -- MixStyle~\cite{zhou2021mixstyle} and Adaptive Risk Minimization (ARM)~\cite{zhang2020arm}. MixStyle~\cite{zhou2021mixstyle} makes use of style transfer, interpolating means and standard deviations in a normalization layer with inputs in a mini-batch coming from different domains. Additionally, to handle distribution shifts at test time, ARM~\cite{zhang2020arm} uses a contextual network that utilizes extra domain information via a meta-learning strategy. For fair comparison, we adopt all baselines to our setup without substantial modification. 
\paragraph{Datasets.}
In our experiments, we use popular DG benchmarks, namely DomainNet~\cite{saito2019semi}, OfficeHome~\cite{venkateswara2017officehome}, PACS~\cite{li2017pacs} and NICO~\cite{he2020nico}. These datasets are ideal candidates for training and evaluating domain generalizable CL methods with multiple domains and a large number of classes.  Specifically, DomainNet is a large-scale dataset containing images of 126 classes from 4 domains: Real, Clipart, Painting and Sketch. OfficeHome is another large-scale benchmark that contains 15K images spanning a total of 65 classes in 4 domains: Real, Clipart, Art and Product. We also consider the PACS dataset which has images from 4 domains: Art, Cartoon, Photo and Sketch. PACS provides challenging recognition scenarios with large domain shifts as described in~\cite{li2017pacs}. Finally, we evaluate on the NICO dataset that has multiple domains called contexts. We consider four domains (Eating, Ground, Water and Grass) from the animal split of the dataset since only these domains contain all classes. Comparatively, we consider smaller task experiments on the PACS and NICO-Animal datasets since the number of categories is limited.
\paragraph{Implementation details.} For all datasets, we follow the provided splits for training and testing. Furthermore, the images are resized to $224 \times 224$. We adopt three cross-domain CL protocols, which consist of 2, 5, and 10 tasks. In our experiments, we exclude one domain for evaluation and consider the remaining ones for training, \eg, for DomainNet, we hold the Clip domain for testing while training on Paint, Real and Sketch samples. 
Following the common practice in CL, some exemplars are also stored in the memory to replay in future iterations. The memory sizes in our experiments are set to 10 for DomainNet and 5 for all other datasets. Note that exemplar selection strategy is not the main focus in this work. Thus, we opt to use random selection and replay the same images in the memory for all methods in our experiments. We train a model for 200 epochs with standard data augmentations (\eg flipping, cropping and color jittering) by using the SGD optimizer with the learning rates of 1$e-$4 for DomainNet and 2$e-$5 for other datasets. 
We use a ResNet-34 model pretrained on the ImageNet as our backbone network. 
As for the distillation loss, we experimentally observed that setting the hyperparameter $\lambda$ to 1$e-$3, 1$e-$3, 1$e-$2, 1$e-$3 works well for LwF~\cite{li2017lwf}, LUCIR~\cite{hou2019lucir}, GeoDL~\cite{simon2021geodl} and our methods, respectively.
The exponential moving average hyperparameter in our method is set to $\gamma=0.96$. As suggested in~\cite{Hinton2015Distillation,li2017lwf}, we set $\tau = 2$ to achieve softer probabilities among
classes. Finally, we found a maximum rank of $r=64$ for the Mahalanobis metric matrices to work well across all protocols and datasets.

\paragraph{Evaluation measurements.}
We assess the baselines and our method for cross-domain CL using two important measurements.  The average accuracy of all tasks  is considered to evaluate the model capability when continually learning new tasks.  Another measurement is the ability to transfer backward from new tasks to old tasks
, which is related to the forgetting rate in cross-domain CL. We follow the backward transfer formulation proposed in~\cite{lopez2017gem}, where  $\mathcal{A}_t$ is the accuracy on task $t$ (\ie, where $y \in {\bigcup}_{i=1}^{t} \mathcal{Y}_i$ from domains $\mathcal{D}_1, \dots, \mathcal{D}_m$).  
Let $\mathcal{A}_t\vert_{\Mat{\Psi}_j}$ be the accuracy for task $t$ evaluated using a model trained from task $1$ to $j$, where $j \leq t$. Then the average accuracy and backward transfer are defined as 
\begin{align}
\begin{split}
 \mathcal{A} = \frac{1}{q}\sum^q_{t=1} \mathcal{A}_t ,\;\;\;\mathcal{B}\mathcal{W} = \frac{1}{q}\sum^q_{t=1}  \mathcal{A}_t\vert_{\Mat{\Psi}_t} - \mathcal{A}_t\vert_{\Mat{\Psi}_q}  
 \;.    
\end{split}
\end{align}
A better model is identified with a larger value of the average accuracy and a lower value of the backward transfer rate.

\begin{table*}[t]
\setlength{\extrarowheight}{6pt}
    \centering
     \resizebox{1\textwidth}{!}{
    \Large\addtolength{\tabcolsep}{-0.2pt}
    \begin{tabular}{c c c c c c  c  c c c c          c c c c c c  c c c c }
    \hline
        \multirow{3}{*}{Method} & & \multicolumn{9}{c}{DomainNet } & & \multicolumn{9}{c}{OfficeHome}\\
        \cline{2-21}
       & &\multicolumn{4}{c}{{10-Task Acc. ($\%\uparrow$)}} & &\multicolumn{4}{c}{{5-Tasks Acc. ($\%\uparrow$)}} &  &\multicolumn{4}{c}{{10-Task Acc. ($\%\uparrow$)}} & &\multicolumn{4}{c}{{5-Tasks Acc. ($\%\uparrow$)}}\\
         & &Clip &Paint &Real &Sketch & &Clip &Paint &Real &Sketch  & &Art &Product &Clipart &Real & &Art &Product &Clipart &Real\\
          \hline
         ERM~\cite{Vapnik1998} & &{60.0}&{51.4}&{60.3}&{53.1} & &{59.7} &50.2 &57.7 &51.7 & &48.8 &52.3 &64.7 &62.4 & &49.7 &51.6 &64.9 &61.9\\
        LwF~\cite{li2017lwf} & &{61.3}&{51.9}&{60.0}&{53.5} & &{62.2} &52.1 &62.6 &54.9 & &49.4 &53.8 &65.2 &63.2 & &49.9 &51.3 &67.5 &63.1\\
        LUCIR~\cite{hou2019lucir} & &{61.1}&{52.1}&{59.7}&{53.0} & &{61.3} &52.7 &61.1 &55.4 & &49.3 &53.6 &65.7 &62.3 & &49.7 &51.6 &67.5 &64.9\\
        GeoDL~\cite{simon2021geodl} & &{61.0}&{50.5}&{58.5}&{54.1} & &{62.1} &52.8 &61.1 &55.5 & &50.6 &53.0 &67.1 &63.1 & &50.5 &52.4 &67.4 &64.2\\
        ARM~\cite{zhang2020arm} & &{57.0}&{49.3}&{62.3}&{51.2} & &{55.4} &51.8 &60.2 &47.7 & &39.8 &55.0 &54.3 &51.7 & &43.6 &56.3 &54.5 &55.4\\
        MixStyle~\cite{zhou2021mixstyle} & &{58.0}&{51.4}&{59.5}&{52.5} & &{59.6} &48.5 &56.0 &53.5 & &47.3 &54.9 &56.3 &56.0 & &48.9 &56.9 &57.7 &59.8\\
        MixStyle + LUCIR & &{62.4}&{50.0}&{59.5}&{52.8} & &58.2 &47.4 &54.8 &51.8 & &51.3 &52.2 &65.1 &62.0 & &49.3 &49.5 &65.5 &63.4\\
       \rowcolor{aliceblue} MSL  (ours)  & &{63.2} &{51.3}&{61.8}&{55.6} & &{63.3} &\textbf{55.4} &63.6 &57.4 & &\textbf{61.6} &61.4 &71.7 &72.7 & &54.3 &\textbf{63.6} &68.0 &67.3\\
       \rowcolor{aliceblue} MSL + Mov (ours)  & &\textbf{63.7} &\textbf{55.0} &\textbf{63.1} &\textbf{56.4} & &\textbf{63.8} &55.3 &\textbf{64.6} &\textbf{58.3} & &61.2 &\textbf{63.0} &\textbf{75.3} &\textbf{73.1} & &\textbf{57.9} &60.2 &\textbf{71.4} &\textbf{70.9}\\
        \hline
    \end{tabular}
    }\caption{Cross-domain continual learning average accuracy numbers for unseen domains with 10-tasks and 5-tasks protocols on the DomainNet~\cite{saito2019semi} and OfficeHome~\cite{venkateswara2017officehome} datasets. } 
    \label{tab:supervised_result_officehome_domainnet}
\end{table*}

\begin{figure*}[!tb]
    \centering
     \subfloat[Clip]{
    \includegraphics[width=0.24\textwidth]{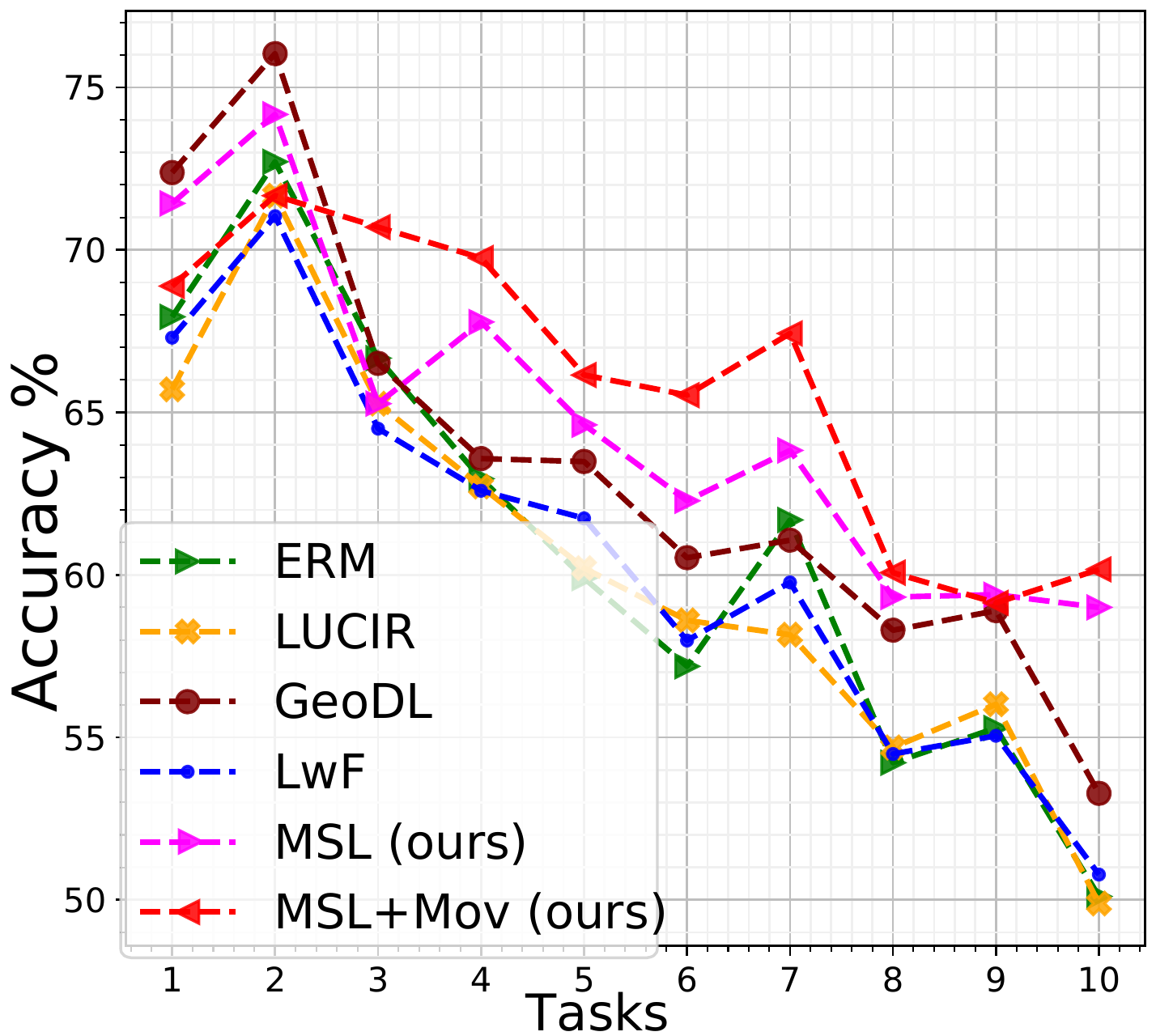}} 
    \subfloat[Paint]{
        \includegraphics[width=0.234\textwidth]{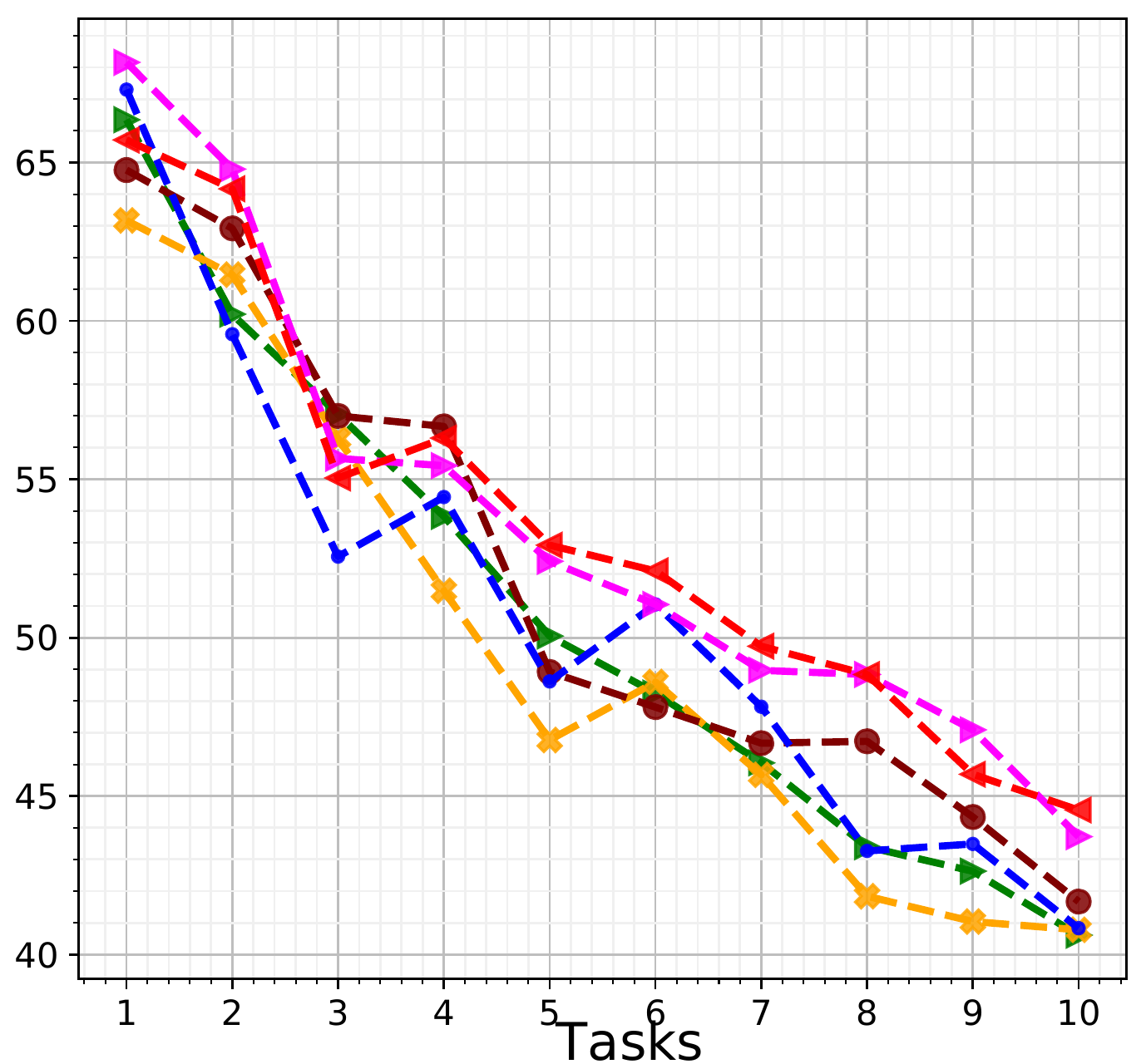}} 
     \subfloat[Real]{
        \includegraphics[width=0.234\textwidth]{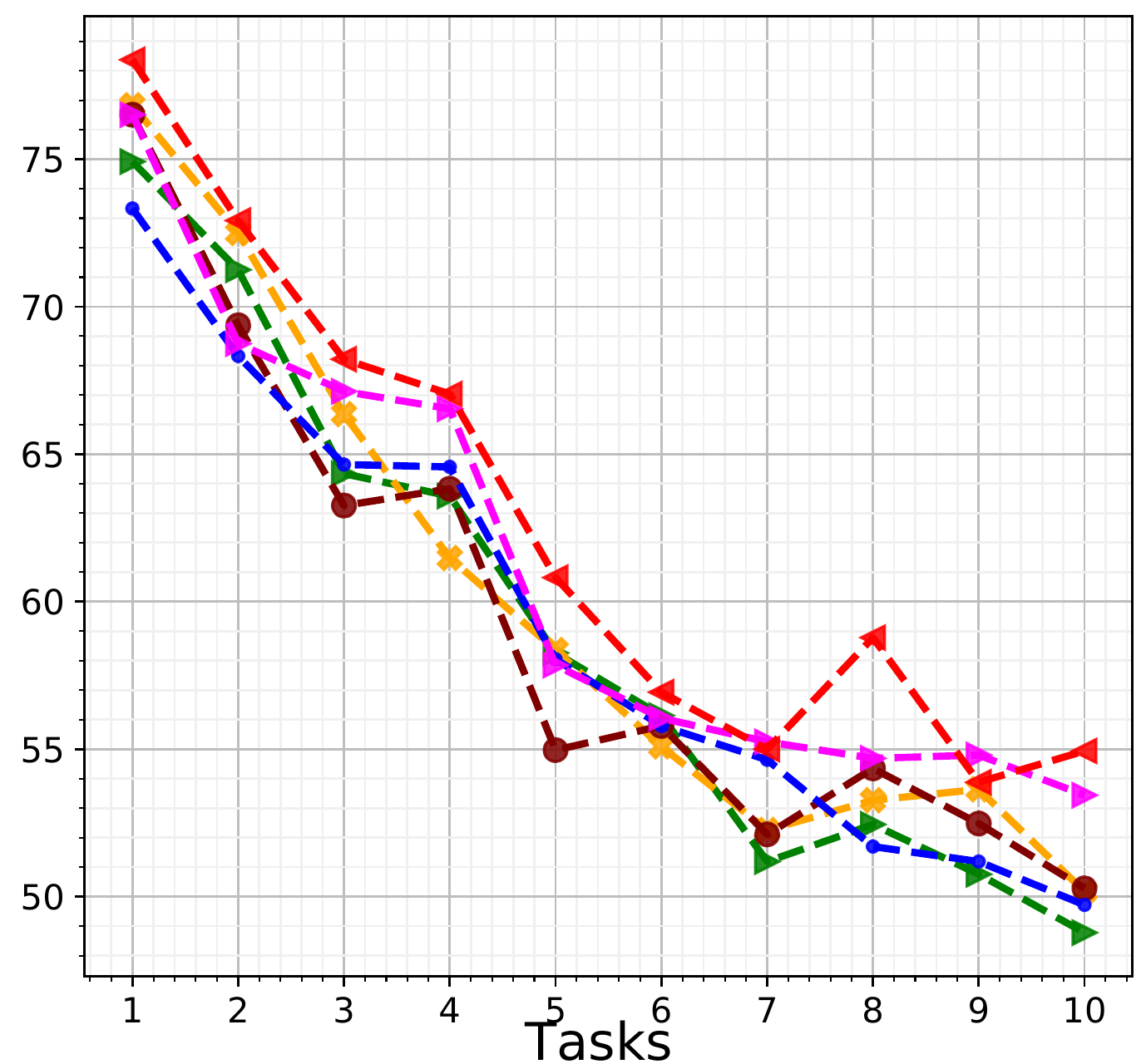}} 
     \subfloat[Sketch]{
        \includegraphics[width=0.24\textwidth]{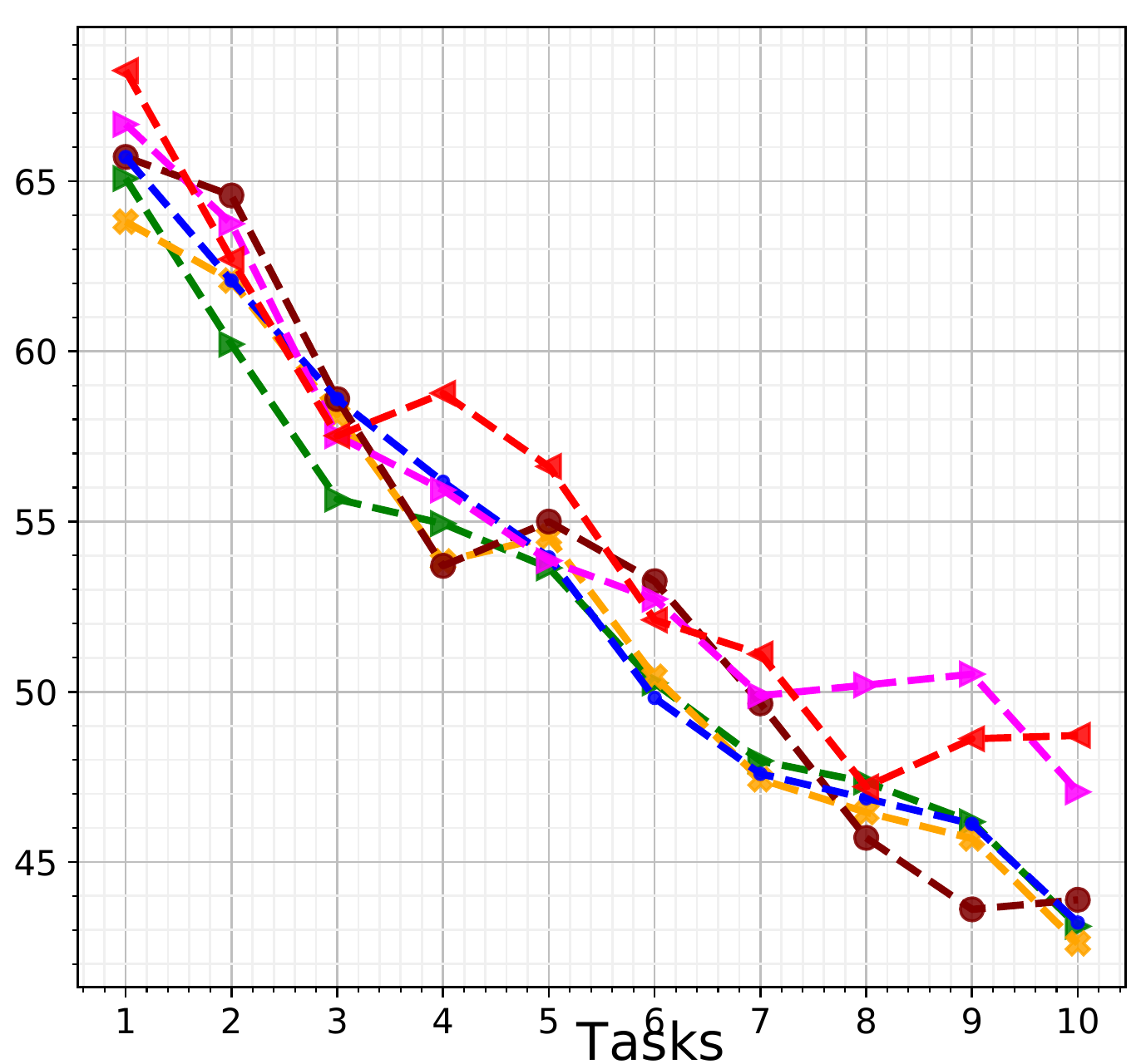}}     
    \caption{The average accuracy numbers of tasks on the unseen domains (Clip, Paint, Real and Sketch) of the DomainNet dataset~\cite{saito2019semi} using the 10-tasks protocol. }%
    \vspace{-0.35cm}
    \label{fig:domainnet_tasks}
\end{figure*}
\begin{figure*}[!tb]
    \centering
     \subfloat[Art]{
    \includegraphics[width=0.24\textwidth]{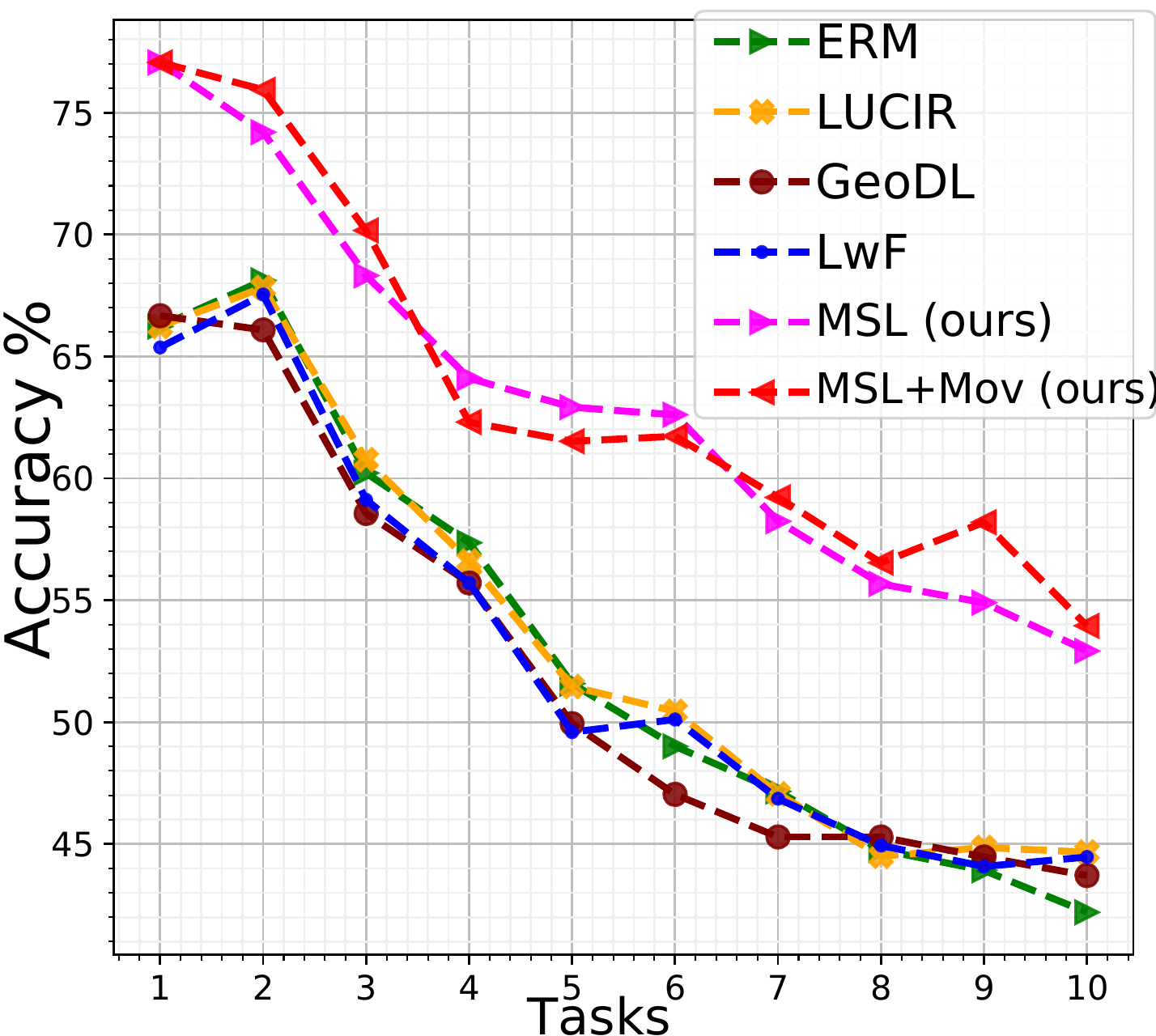}} 
    \subfloat[Product]{
        \includegraphics[width=0.234\textwidth]{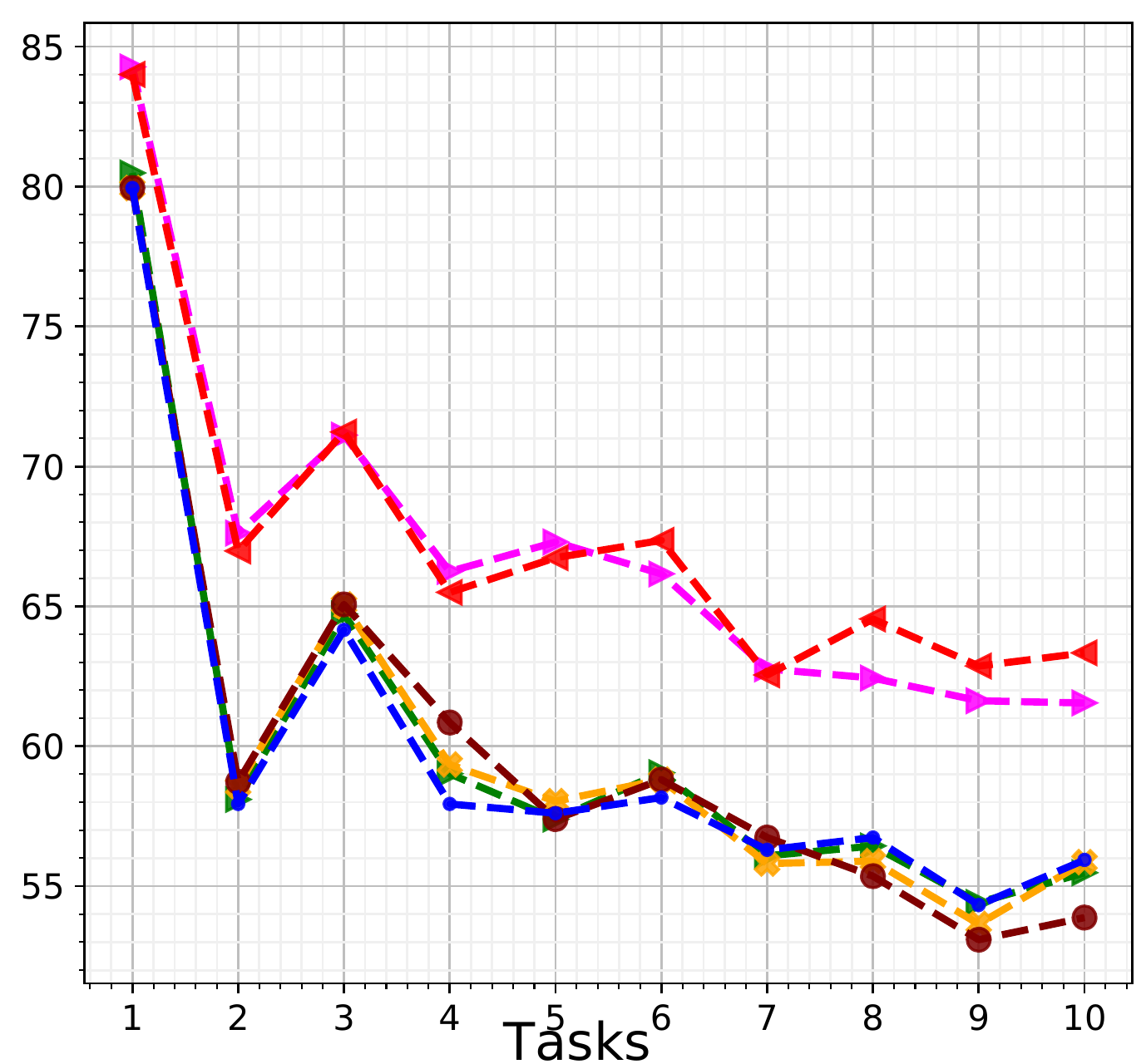}} 
     \subfloat[Clipart]{
        \includegraphics[width=0.234\textwidth]{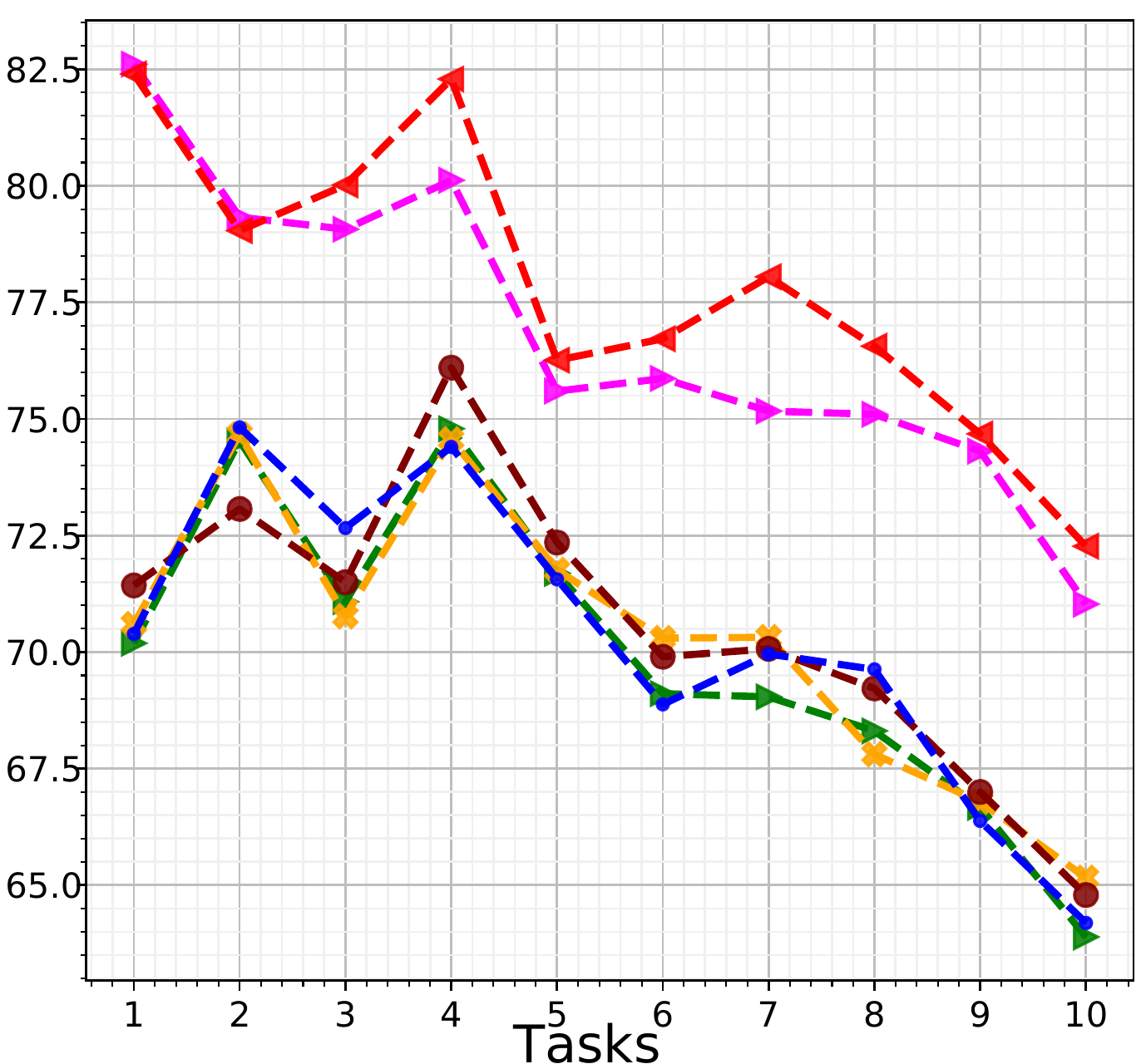}} 
     \subfloat[Real]{
        \includegraphics[width=0.24\textwidth]{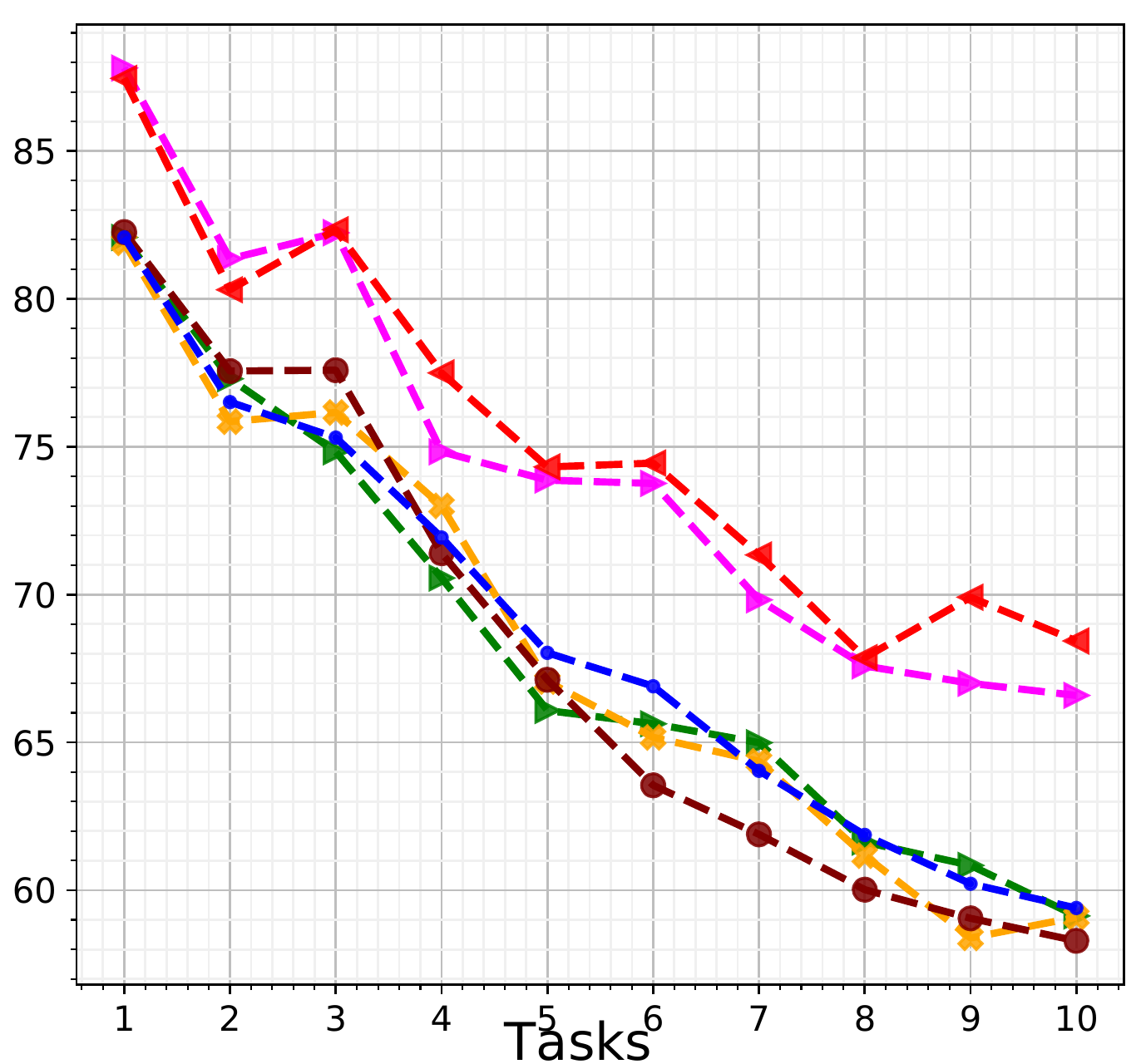}}     
    \caption{The average accuracy numbers of tasks on the unseen domains (Art, Product, Clipart and Real) of the OfficeHome dataset~\cite{venkateswara2017officehome} using the 10-tasks protocol. 
    }
    \vspace{-0.35cm}
    \label{fig:officehome_tasks}
\end{figure*}

\subsection{Supervised Cross-Domain Continual Learning}

We evaluate our cross-domain CL methods in supervised fashions when available classes from a benchmark are split into 5-tasks and 10-tasks. 
As shown in Table~\ref{tab:supervised_result_officehome_domainnet},  each column corresponds to the performances when samples of a single domain are entirely excluded (considered unseen) from the training. In addition, we report results of the accuracy numbers on the seen domains in our supplementary material.
As can be seen in the Table, both our methods with Mahalanobis metrics and biases (MSL and MSL + Mov) comfortably outperform all the competitors in CL and DG. Here, MSL with knowledge distillation and exponential moving average (MSL + Mov) improves over MSL and achieves the highest accuracy. To mention one instance, when generalizing to DomainNet-Sketch as an unseen domain, our MSL + Mov obtains 56.4\%, which is 2.3\% higher compared to GeoDL's performance of 54.1\% for the 10-tasks protocol. This improvement trend also appears on the seen domains. 

Furthermore, in Fig.~\ref{fig:domainnet_tasks} and Fig.~\ref{fig:officehome_tasks}, we show the average classification accuracy numbers of tasks obtained by our methods in comparison to prior CL methods on the DomainNet and OfficeHome datasets for the 10-tasks protocol. Overall, our methods outperform the baselines, with more significant performance gaps on the OfficeHome dataset. 
In addition, we evaluate on the PACS and the NICO-Animal datasets. The results are  shown in Table~\ref{tab:supervised_result_pacs_nico}.
Here, the average accuracy of our methods is superior to other competitors by at least 2\% margin. MSL + Mov clearly shows the benefits of knowledge distillation to a history of models, with a gap of 2.1\% over MSL in the best case.

\begin{figure}[!tb]
    \centering
     \subfloat[DomainNet ($\%\downarrow$)]{
    \includegraphics[width=0.24\textwidth]{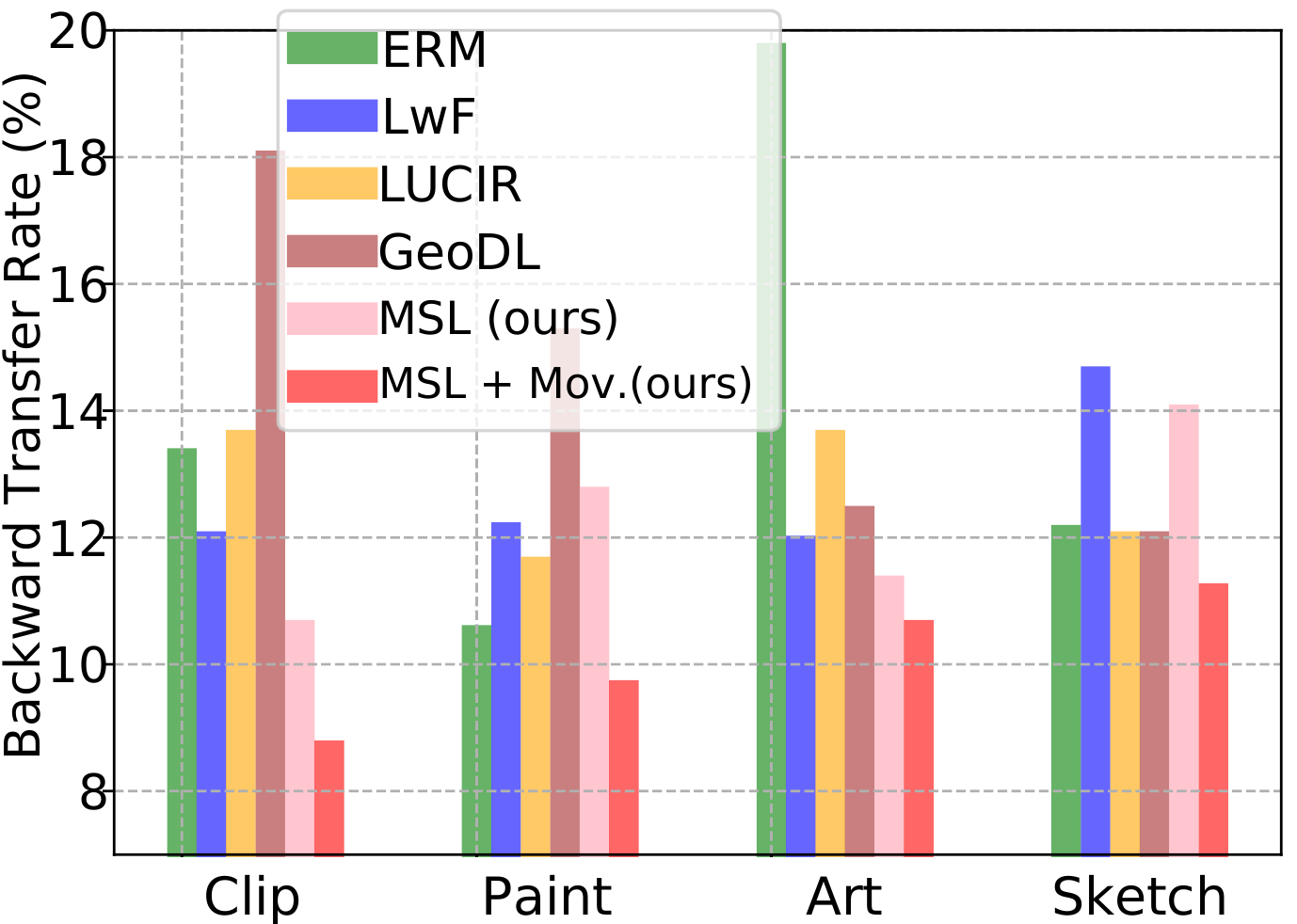}} 
    \subfloat[OfficeHome ($\%\downarrow$)]{
        \includegraphics[width=0.237\textwidth]{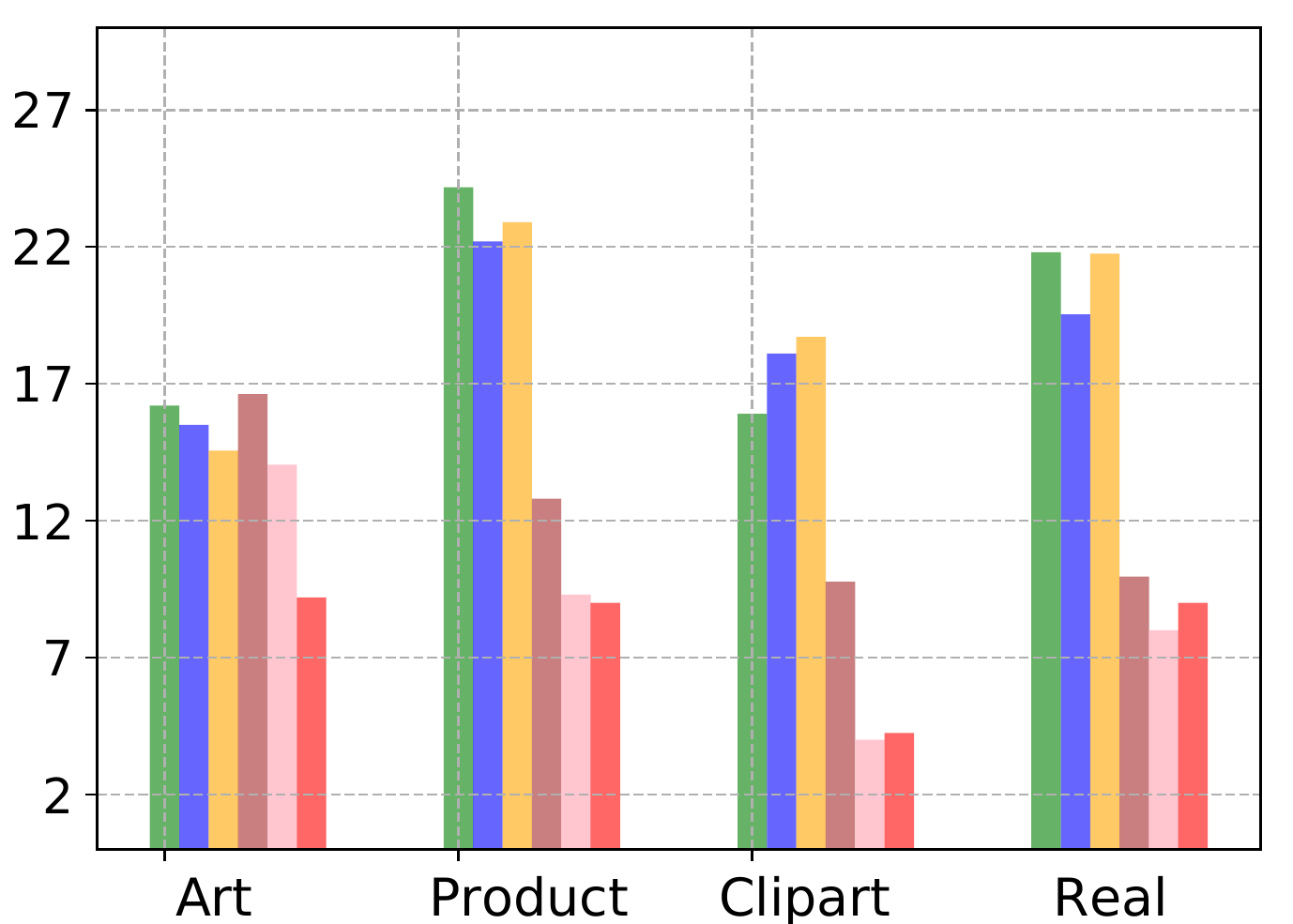}} 
        \caption{The backward transfer performance (absolute values) on DomainNet~\cite{saito2019semi} and OfficeHome~\cite{venkateswara2017officehome} datasets (the lower the better) for the 10-tasks protocol.}
        \label{fig:backward_transfer}
\end{figure}

Overall, we observe that standard CL algorithms largely fail to prevent {catastrophic forgetting} in the cross-domain setting, indicated by high backward transfer rates shown in Fig.~\ref{fig:backward_transfer}. In contrast, on average, our method with exponential moving average, MSL + Mov, can achieve the lowest backward transfer rate of 10.1\% on the DomainNet and 7.8\% on the OfficeHome datasets for the 10-tasks protocol. The closest competitors to ours report 12.8\% (by LUCIR) and 12.3\% (by GeoDL), respectively.  Note that the method with the lower backward transfer rate is better. 

\begin{table}[t]
  \setlength{\extrarowheight}{6pt}
    \centering
    \resizebox{0.49\textwidth}{!}{
    \Large\addtolength{\tabcolsep}{-2pt}
    \begin{tabular}{c  c c c c c c c c c   }
    \hline
   \multirow{2}{*}{Method}   & \multicolumn{4}{c}{NICO-Animal ($\%\uparrow$)  } &  &\multicolumn{4}{c}{PACS ($\%\uparrow$) }\\
         &Eating &Ground &Water &Grass &  &Art &Cartoon &Photo &Sketch  \\
        \hline
        ERM~\cite{Vapnik1998} &88.0 &86.5 &82.3 &84.3 & &76.3 &82.9 &84.7 &61.9 \\
        LwF~\cite{li2017lwf} &88.2 &86.2 &83.3 &84.3 & &76.4 &82.4 &85.5 &62.4\\
        LUCIR~\cite{hou2019lucir} &88.1 &86.6 &83.3 &84.3 & &76.5 &82.1 &84.2 &62.2\\
        GeoDL~\cite{simon2021geodl} &87.9 &86.1 &82.5 &83.3 & &72.8 &83.6 &85.4 &60.5\\
        ARM~\cite{zhang2020arm}   &86.2	&83.5	&80.8	&83.5 & &65.1	&83.3	&84.9	&64.9\\
        MixStyle~\cite{zhou2021mixstyle}  &86.1 &84.0 &81.0 &83.4 & &73.3 &82.6 &81.1 &63.5\\
        MixStyle + LUCIR  &88.0 &82.6 &81.9 &83.0 & &70.3 &82.7 &83.6 &63.6\\
        \rowcolor{aliceblue} MSL (ours) &89.9 &86.2 &84.4 &85.2 & &\textbf{77.3} &82.1 &87.0 &62.8\\
       \rowcolor{aliceblue} MSL + Mov (ours) &\textbf{91.3} &\textbf{87.9} &\textbf{85.0} &\textbf{87.2} & &77.2 &\textbf{84.1} &\textbf{89.0} &\textbf{64.9} \\
        \hline
    \end{tabular}
    }\caption{Domain generalization test for 2-tasks supervised learning on NICO-Animal~\cite{he2020nico} and PACS~\cite{li2017pacs} datasets.} 
    \label{tab:supervised_result_pacs_nico}
\end{table}     
\subsection{Ablation Studies and Analyses}
We investigate how hyperparameters impact the performance of our proposal. Below, we show the impact of matrix rank for the Mahalanobis similarity learning and how the memory size affects the performance. 
\begin{figure}[h]
    \centering
     \subfloat[MSL]{
    \includegraphics[width=0.23\textwidth]{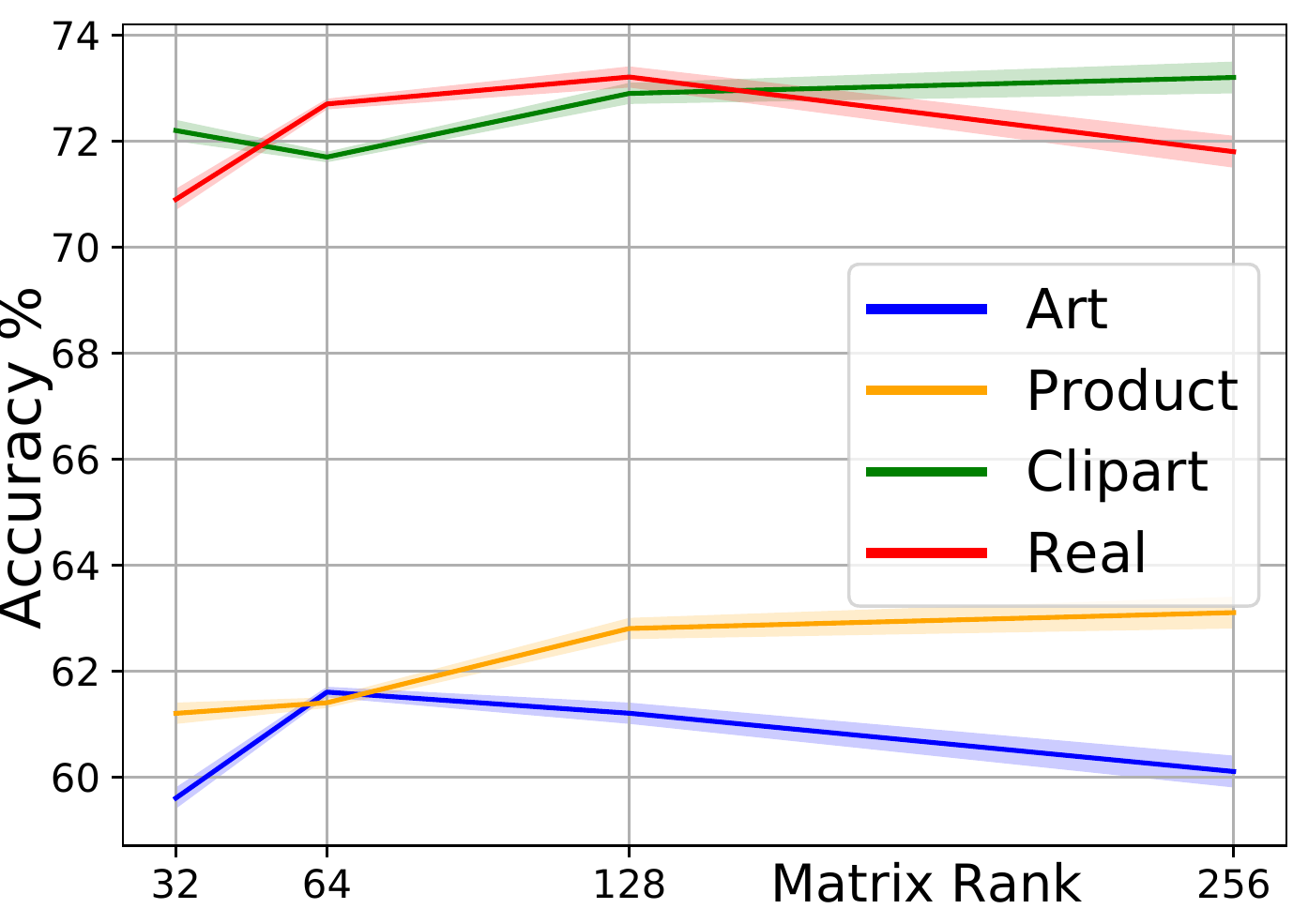}} 
    \subfloat[MSL + Mov]{
        \includegraphics[width=0.23\textwidth]{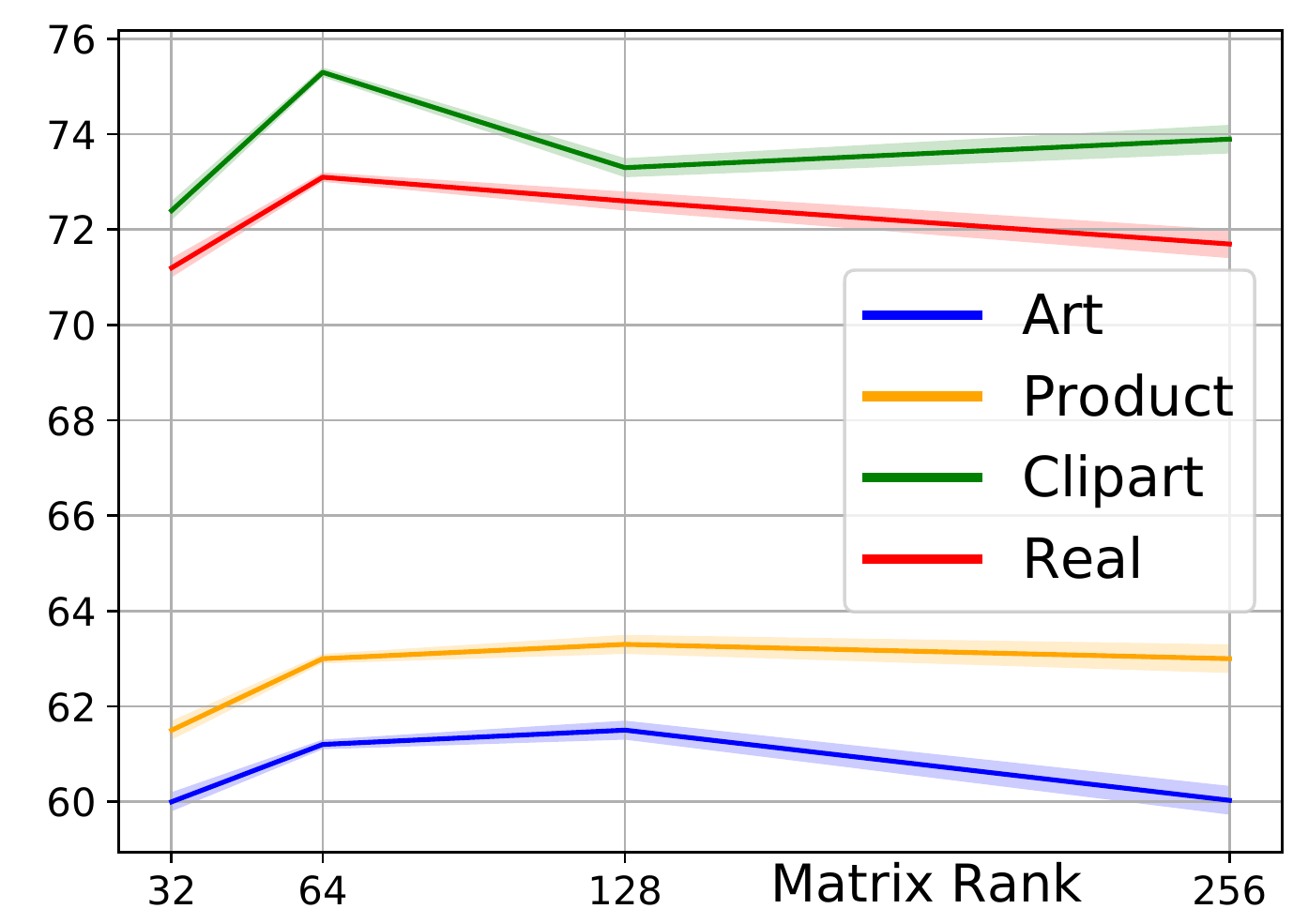}} 
        \caption{The impact of varying $r$ in learning the Mahalanobis metric matrices on the average accuracy  using the unseen domains of the Officehome dataset~\cite{venkateswara2017officehome} and the 10-tasks protocol.}
        \label{fig:matrix_rank}
\end{figure}
\begin{figure}[h]
    \centering
     \subfloat[10 memory size ($\%\uparrow$)]{
    \includegraphics[width=0.235\textwidth]{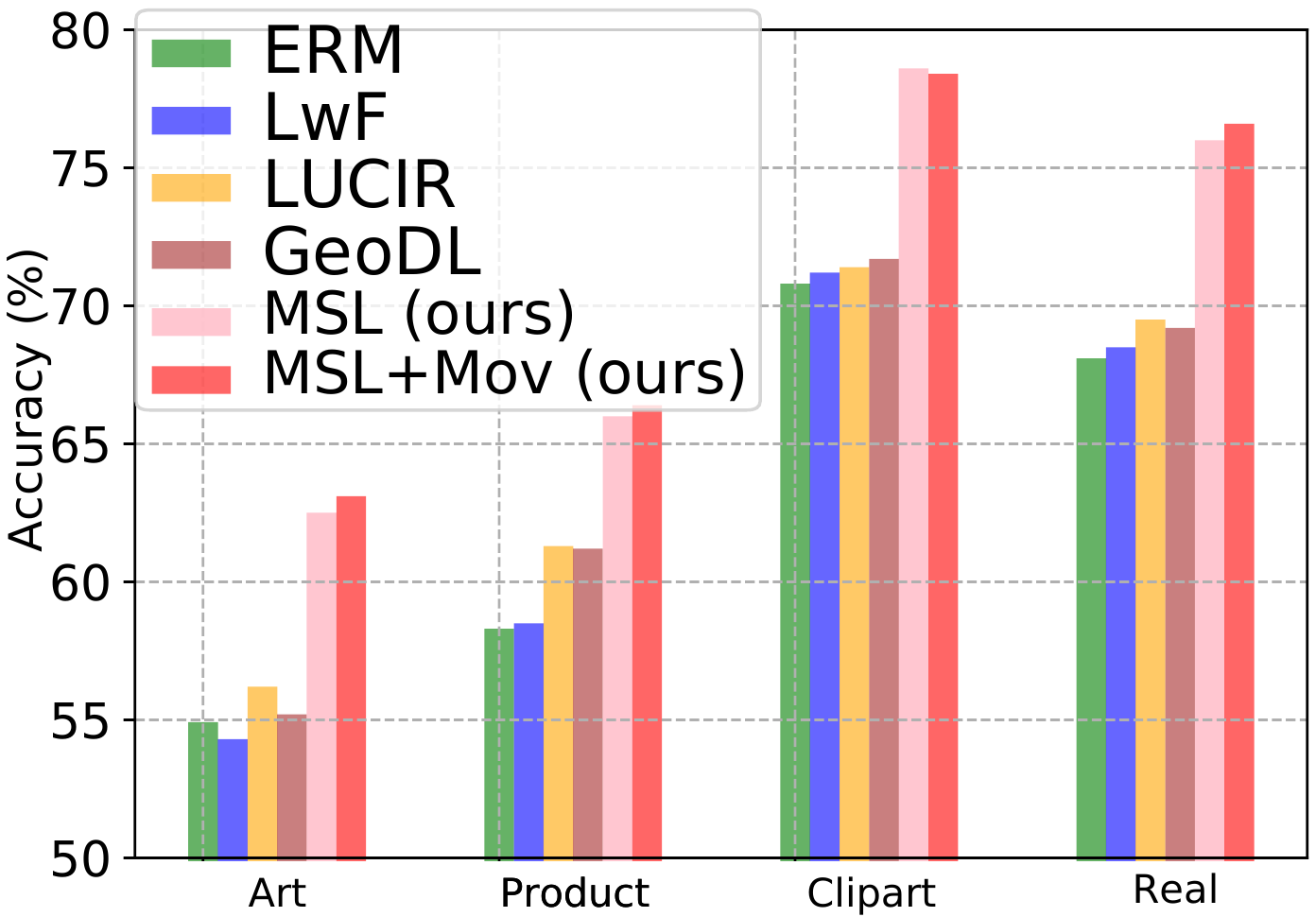}} 
    \subfloat[20 memory size ($\%\uparrow$)]{
        \includegraphics[width=0.232\textwidth]{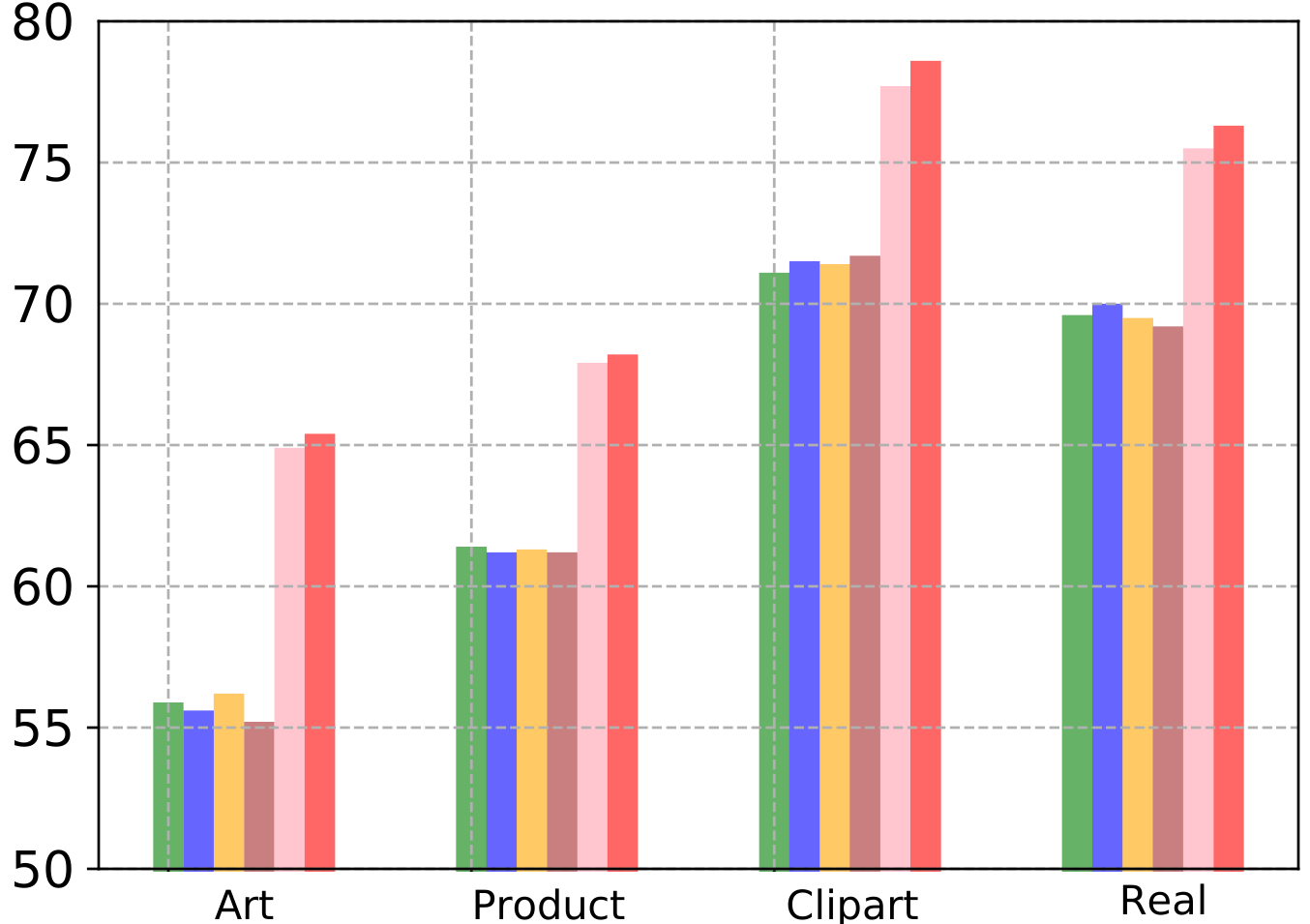}} 
        \caption{The impact of varying memory size on the average accuracy using the unseen domains of the Officehome dataset~\cite{venkateswara2017officehome} and the 10-tasks protocol.}
        \label{fig:memory_ablation}
\end{figure}

\paragraph{Impact of varying the maximum rank.}
Our approach employs a low-rank strategy for the metric matrices. We investigate the performances when four different $r$ values (\eg, 32, 64, 128 and 256) are used. We observe in Fig.~\ref{fig:matrix_rank} that varying $r$  would change the accuracy by less than 1.5\%. The plots also  show that setting a large value for $r$, which translates to having more parameters, does not directly increase the performance.  As a rule of thumb, the matrix maximum rank value that we use for all experiments is set to 64. 
\paragraph{Impact of varying the memory size.}
Below, we investigate how increasing the memory size impacts the average accuracy over the 10-tasks protocol with 10 and 20 exemplars. Fig.~\ref{fig:memory_ablation} shows that having more exemplars in the memory attains higher accuracy for all methods. Our proposed approach can still benefit from more exemplars and outperform baselines. We observe that our method leads by over 5\% for both 10 and 20 exemplar sizes in the memory.

\section{Limitations and Conclusions}
\label{sec5:Conc}

Our setup and method can be limited from two aspects. First, like many other continual learning setups, one underlying assumption of our setup is that some exemplars are stored for replay. This might restrict its use in some applications with strict privacy regulations. Second, the number of parameters in our methods grow linearly following the number of tasks. Though, we have proposed a memory efficient alternative that scales well to many applications, this might still limit the practical use when very large-scale applications with severe memory constraints are desired. %

\vspace{2mm}
We propose an approach to generalize across training domains while mitigating the so-called \textit{catastrophic forgetting} issue via Mahalanobis similarity learning and knowledge distillation with exponential moving average update. In our evaluation, we follow the so-called leave-one-domain-out protocol where a test domain is not seen during training. As our experimental evaluations indicate, our methods comfortably outperform the existing methods in both class continual learning  and domain generalization on challenging datasets, namely DomainNet, OfficeHome, PACS and NICO-Animal. The ablation studies also show that our method consistently improves over the baselines in various conditions and has low sensitivity to the choice of hyperparameters.

{\small
\bibliographystyle{ieee_fullname}
\bibliography{egbib}
}

 \newpage  
\makeatletter
\newcommand*{\addFileDependency}[1]{
  \typeout{(#1)}
  \@addtofilelist{#1}
  \IfFileExists{#1}{}{\typeout{No file #1.}}
}
\makeatother

\newcommand*{\myexternaldocument}[1]{
    \externaldocument{#1}%
    \addFileDependency{#1.tex}%
    \addFileDependency{#1.aux}%
}


\renewcommand{\thefigure}{A\arabic{figure}}
\renewcommand{\thetable}{A\arabic{table}}
\renewcommand{\thesection}{\Alph{section}}


\noindent\textbf{\large Supplementary Material}

While the main focus of the supervised experiments is evaluation on the unseen data in leave-one-domain-out protocol, in this supplementary material, we provide additional results on the test sets of the seen domains. 

\section{Additional Results on the Seen Domains}
In the main paper, we have shown that our proposed approach is superior to all the competitors when evaluating on the unseen domains in the supervised experiments. Here, we provide additional results on seen domains to demonstrate that our methods achieve improvements on all domains, including the seen ones. As we can observe in Table~\ref{tab:seendomains_domainnet_officehome}, our MSL and MSL + Mov are also superior to other baselines when seen domains are considered. For example, the best accuracy number of 71.0\% on DomainNet-Real is obtained by MSL + Mov (see the last row in Table~\ref{tab:seendomains_domainnet_officehome}) with the gap of 4.3\% compared to the recent method of GeoDL. 

Furthermore, Table~\ref{tab:seendomain_pacs_nico} shows that our methods can achieve the best accuracy numbers on the unseen domains of the PACS and NICO datasets. On the seen domains, we observe that the best accuracy number is sometimes about 1-2\% better than ours (\eg on the NICO dataset row-col: water-ground and grass-ground). However, our proposal achieves the best average accuracy over both the test sets of seen and unseen domains.

\begin{table*}[!tb]
    \centering
     \resizebox{0.9\textwidth}{!}{
    \Large\addtolength{\tabcolsep}{-0.2pt}
    \begin{tabular}{c c c c c c  c  c c c c         c c c c c c c  c c c c }
    \hline
        \multirow{3}{*}{Method} & \multicolumn{10}{c}{DomainNet } & & \multicolumn{10}{c}{OfficeHome}\\
        \cline{2-22}
        &\multirow{2}{*}{Unseen Domain} &\multicolumn{4}{c}{{10-Task Acc. ($\%\uparrow$)}} & &\multicolumn{4}{c}{{5-Tasks Acc. ($\%\uparrow$)}} & &\multirow{2}{*}{Unseen Domain} &\multicolumn{4}{c}{{10-Task Acc. ($\%\uparrow$)}} & &\multicolumn{4}{c}{{5-Tasks Acc. ($\%\uparrow$)}}\\
         & &Clip &Paint &Real &Sketch & &Clip &Paint &Real &Sketch & & &Art &Product &Clipart &Real & &Art &Product &Clipart &Real\\
        \hline
        \multirow{4}{*}{ERM~\cite{Vapnik1998}} &Clip &{\cellcolor{gray!30}60.0}&{56.7}&{65.4}&{58.7} & &{\cellcolor{gray!30}}59.7 &52.7 &62.0 &54.0 & &Art &\cellcolor{gray!30}48.8	&55.8	&69.1 &67.1 &  &\cellcolor{gray!30}49.7 &56.1 &70.2 &67.9\\ 
        &Paint &{64.4}&{\cellcolor{gray!30}51.4}&{65.3}&{55.7} & &63.6 &\cellcolor{gray!30}50.2 &61.7 &55.2 & 
        &Product &52.7	&\cellcolor{gray!30}52.3	&68.9	&67.0 & &54.9 &\cellcolor{gray!30}51.6 &64.9 &61.9\\  
        &Real &{64.6}&{53.8}&{\cellcolor{gray!30}60.3}&{55.9} & &61.3	&50.2	&\cellcolor{gray!30}57.7	&54.4 & &Clipart &52.3 &57.0	&\cellcolor{gray!30}64.7 &66.1 & &53.8	&55.6	&\cellcolor{gray!30}64.9	&67.2\\  
        &Sketch &{65.2}&{56.0}&{65.4}&{\cellcolor{gray!30}53.1} & &61.5	&52.9	&63.0	&\cellcolor{gray!30}51.7 & &Real &52.3	&54.8	&66.4	&\cellcolor{gray!30}62.4 & &51.9	&55.4	&66.6	&\cellcolor{gray!30}61.9\\ 
        \hline
         \multirow{4}{*}{LwF~\cite{li2017lwf}}  &Clip &\cellcolor{gray!30}61.3	&57.0 &67.4	&58.0  & &\cellcolor{gray!30}62.2	&56.8	&67.6	&60.3   &&Art &\cellcolor{gray!30}49.4	&57.5	&69.2	&67.1 &  &\cellcolor{gray!30}49.9	&57.8	&70.8	&68.2\\
        &Paint &64.7	&\cellcolor{gray!30}51.9	&65.4	&56.4 & &65.2	&\cellcolor{gray!30}52.1	&67.0	&60.4 &&Product &53.7	&\cellcolor{gray!30}53.8	&70.5	&68.1 & &54.8	&\cellcolor{gray!30}51.3	&70.7	&68.6\\
        &Real &65.0	&54.7	&\cellcolor{gray!30}60.0 &56.1 & &64.89	&52.2	&\cellcolor{gray!30}62.6	&59.2 & &Clipart &52.6	&57.4	&\cellcolor{gray!30}65.2 &66.4 & &54.2	&56.4	&\cellcolor{gray!30}67.5	&68.5 \\
        &Sketch &65.8	&56.2	&66.7	&\cellcolor{gray!30}53.5 & &65.0	&54.8	&66.3	&\cellcolor{gray!30}54.9 &&Real &53.1	&55.7	&66.1	&\cellcolor{gray!30}63.2 & &53.1  &55.7  &66.1      &\cellcolor{gray!30}63.1\\
          \hline
        \multirow{4}{*}{LUCIR~\cite{hou2019lucir}}  &Clip &\cellcolor{gray!30}61.1	&57.1	&65.7 &57.9 & &\cellcolor{gray!30}61.3	&56.5	&67.1	&60.3 &&Art &\cellcolor{gray!30}49.3	&56.2	&69.0	&67.2 &    &\cellcolor{gray!30}49.7	&57.7	&70.9	&68.5\\
        &Paint &65.0 &\cellcolor{gray!30}52.1	&65.5	&55.7 &  &65.1	&\cellcolor{gray!30}52.7	&66.3	&59.3 &&Product &53.4	&\cellcolor{gray!30}53.6	&71.0	&67.9 &    &55.0	&\cellcolor{gray!30}51.6	&70.3	&68.3\\
        &Real &64.9	&54.7	&\cellcolor{gray!30}59.7	&56.8  & &64.0	&52.7	&\cellcolor{gray!30}61.1	&59.2  & &Clipart &52.8	&56.4	&\cellcolor{gray!30}65.7	&66.3 &  &54.3	&56.5	&\cellcolor{gray!30}67.5	&68.6\\
        &Sketch &66.1	&56.0	&65.4	&\cellcolor{gray!30}53.0 & &64.9	&55.2	&66.4	&\cellcolor{gray!30}55.4 &&Real &52.8	&55.0	&66.4	&\cellcolor{gray!30}62.3 & &53.7	&56.4	&67.2	&\cellcolor{gray!30}64.9\\
          \hline
        \multirow{4}{*}{GeoDL~\cite{simon2021geodl}} 
        &Clip &\cellcolor{gray!30}61.0	&55.6	&64.9	&58.2 & &\cellcolor{gray!30}62.1	&56.7	&67.3	&60.1 &&Art &\cellcolor{gray!30}50.6	&56.7	&69.6	&68.0 & &\cellcolor{gray!30}50.5	&57.6	&71.0	&68.9\\
        &Paint &65.6	&\cellcolor{gray!30}50.5	&64.4	&56.2 & &64.9	&\cellcolor{gray!30}52.8	&66.8	&59.6 &&Product &51.9	&\cellcolor{gray!30}53.0	&69.7	&67.2 & &54.6	&\cellcolor{gray!30}52.4	&70.4	&68.4\\
        &Real &64.7	&53.1	&\cellcolor{gray!30}58.5	&57.5 & &63.8	&53.0	&\cellcolor{gray!30}61.1	&59.1    &&Clipart &54.5	&58.4	&\cellcolor{gray!30}67.1	&68.3 & &54.4	&56.5	&\cellcolor{gray!30}67.4	&68.7\\
        &Sketch &65.8	&56.0	&66.7	&\cellcolor{gray!30}54.1 & &65.1	&55.1	&66.2	&\cellcolor{gray!30}55.5  &&Real &52.5	&55.9	&67.0	&\cellcolor{gray!30}63.1 & &53.6	&55.7	&66.8	&\cellcolor{gray!30}64.2\\
          \hline
          \multirow{4}{*}{ARM~\cite{zhang2020arm}} &Clip &\cellcolor{gray!30}57.0	&50.9	&65.0	&50.5  & &\cellcolor{gray!30}55.4	&52.2	&64.5	&49.9 & &Art &\cellcolor{gray!30}39.8	&56.2	&58.2	&51.7 & &\cellcolor{gray!30}43.6	&58.4	&61.7	&57.0 \\
        &Paint &61.1	&\cellcolor{gray!30}49.3	&64.0	&52.9 & &62.7	&\cellcolor{gray!30}51.8	&66.3	&54.4 &&Product &46.4	&\cellcolor{gray!30}55.0	&60.3	&54.8 & &50.0	&\cellcolor{gray!30}56.3	&62.8	&58.3\\
        &Real &60.4	&51.7	&\cellcolor{gray!30}62.3	&52.9  & &59.2	&51.6	&\cellcolor{gray!30}60.2	&53.0    &&Clipart &45.3	&55.4	&\cellcolor{gray!30}54.3	&53.0 & &48.7	&56.9	&\cellcolor{gray!30}54.5	&54.8 \\
        &Sketch &61.9	&52.5	&65.8	&\cellcolor{gray!30}51.2 & &60.2	&53.0	&64.6	&\cellcolor{gray!30}47.7 &&Real &43.6	&54.7	&56.6	&\cellcolor{gray!30}51.7 & &48.7	&56.8	&60.4	&\cellcolor{gray!30}55.4\\
          \hline
          \multirow{4}{*}{MixStyle~\cite{zhou2021mixstyle}} &Clip &\cellcolor{gray!30}58.0	&53.4	&65.7	&52.7 & &\cellcolor{gray!30}59.6	&56.1	&69.8	&55.5 &&Art &\cellcolor{gray!30}47.3	&60.7	&65.1	&60.5 & &\cellcolor{gray!30}48.9	&60.9	&65.8	&61.3\\
        &Paint &60.7	&\cellcolor{gray!30}51.4	&65.6	&52.3 & &65.5	&\cellcolor{gray!30}48.5	&69.4	&57.0 &&Product &48.6	&\cellcolor{gray!30}54.9	&65.1	&61.0 & &51.2	&\cellcolor{gray!30}56.9	&68.1	&65.2 \\
        &Real &60.8	&53.4	&\cellcolor{gray!30}59.5	&54.6 & &63.6	&57.3	&\cellcolor{gray!30}56.0	&57.2 &&Clipart &48.5	&58.9	&\cellcolor{gray!30}56.3	&57.3 & &50.0	&61.6	&\cellcolor{gray!30}57.7	&58.7\\
        &Sketch &62.3	&53.7	&66.8	&\cellcolor{gray!30}52.5 & &63.8	&57.6	&69.2	&\cellcolor{gray!30}53.5 &&Real &48.2	&56.7	&62.5	&\cellcolor{gray!30}56.0 & &49.4	&59.3	&65.3	&\cellcolor{gray!30}59.8\\
          \hline
        \rowcolor{aliceblue}  &Clip &\cellcolor{gray!30}63.2	&56.6	&68.1 &59.1 & &\cellcolor{gray!30}63.3	&58.3	&70.4	&61.5 &&Art &\cellcolor{gray!30}\textbf{61.6}	&64.1	&75.5	&74.5 & &\cellcolor{gray!30}54.3	&64.9	&72.6	&68.0\\
        \rowcolor{aliceblue} &Paint &67.1 	&\cellcolor{gray!30}51.3	&67.2	&58.3 & &65.8	&\cellcolor{gray!30}\textbf{55.4}	&69.1	&60.9 &&Product &61.5	&\cellcolor{gray!30}61.4	&77.7	&74.6 & &59.1	&\cellcolor{gray!30}\textbf{63.6}	&73.1	&69.9\\
        \rowcolor{aliceblue}  &Real &67.2	&55.5	&\cellcolor{gray!30}61.8	&57.5  & & 64.7	&54.2	&\cellcolor{gray!30}63.6	&60.0  &&Clipart &60.4	&63.0	&\cellcolor{gray!30}71.7	&73.1 & &60.3	&64.6	&\cellcolor{gray!30}68.0	&67.5\\
        \rowcolor{aliceblue} \multirow{-4}{*}{{MSL} (ours)}  &Sketch &67.8	&57.7	&67.9	&\cellcolor{gray!30}55.6 & &65.4	&57.1	&68.6	&\cellcolor{gray!30}57.4 &&Real &61.5	&64.4	&74.9	&\cellcolor{gray!30}72.7 & &58.8	&63.6	&70.6	&\cellcolor{gray!30}67.3\\
          \hline
         \rowcolor{aliceblue}  &Clip &\cellcolor{gray!30}\textbf{63.7}	&58.2	&71.0	&60.9 & &\cellcolor{gray!30}\textbf{63.8}	&58.9	&71.1	&62.6 &&Art &\cellcolor{gray!30}61.2	&65.7	&76.8	&75.1 & &\cellcolor{gray!30}\textbf{57.9}	&63.4	&75.3	&74.0\\
        \rowcolor{aliceblue}  &Paint &68.0	&\cellcolor{gray!30}\textbf{55.0}	&69.5	&58.7 & &66.4	&\cellcolor{gray!30}55.3	&69.7	&61.3 &&Product &63.3	&\cellcolor{gray!30}\textbf{63.0}	&76.4	&75.2 & &60.3	&\cellcolor{gray!30}60.2	&75.1	&74.9 \\
        \rowcolor{aliceblue}  &Real &67.7	&57.7	&\cellcolor{gray!30}\textbf{63.1}	&59.8 & &65.7	&55.2	&\cellcolor{gray!30}\textbf{64.6}	&61.4  & &Clipart &63.7	&64.3	&\cellcolor{gray!30}\textbf{75.3}	&73.7 & &60.3	&61.8	&\cellcolor{gray!30}\textbf{71.4}	&73.2\\
        \rowcolor{aliceblue}  \multirow{-4}{*}{{MSL + Mov.} (ours)}  &Sketch &68.1	&60.6	&71.0	&\cellcolor{gray!30}\textbf{56.4} & &66.6	&58.0	&69.3	&\cellcolor{gray!30}\textbf{58.3}  &&Real &61.3	&64.9	&73.7	&\cellcolor{gray!30}\textbf{73.1} & &56.6	&62.5	&73.0	&\cellcolor{gray!30}\textbf{70.9}\\
         \hline
    \end{tabular}
    }
    \caption{Cross-domain continual learning average accuracy with 10-tasks and 5-tasks protocols, and 10 exemplar size and 5 exemplar size in the memory on DomainNet and OfficeHome datasets, respectively. Within each block that corresponds to a method, the domain shown in the beginning of each row is excluded from training, while each column shows the achieved accuracy when a test set of a domain is used. Therefore, the grey cells show the accuracy on unseen domains in leave-one-domain-out protocol while the off-diagonal cells are the accuracy numbers on the test set of seen domains. }
    \label{tab:seendomains_domainnet_officehome}
\end{table*}

\begin{table*}[!ht]
 \centering
     \resizebox{0.6\textwidth}{!}{
    \Large\addtolength{\tabcolsep}{-0.2pt}
    \begin{tabular}{c  c  c c c c  c  c  c c c c }
    \hline
 \multirow{2}{*}{Method} & &\multicolumn{4}{c}{PACS} & &  &\multicolumn{4}{c}{NICO}\\
 \cline{2-12}
 & Unseen Domain &Art &Cartoon &Photo &Sketch & &Unseen Domain &Eating &Ground &Water &Grass \\
      \hline
\multicolumn{1}{c}{\multirow{4}{*}{ERM~\cite{Vapnik1998}}} &Art
  &\cellcolor{gray!30}88.0 &
  86.4 &
  84.0 &
  83.8 &
   &Eating
   & 
  \cellcolor{gray!30}76.3 &
  85.1 &
  89.1 &
  69.8 \\
\multicolumn{1}{c}{} &Cartoon & 89.5 & \cellcolor{gray!30}86.7 & 86.2 & 85.3 &  &Ground  & 76.6 & \cellcolor{gray!30}82.9 & 85.3 & 68.0 \\
\multicolumn{1}{c}{} &Photo & 87.3 & 86.1 & \cellcolor{gray!30}82.3 & 85.1 &  &Water  & 74.1  & 81.2 & \cellcolor{gray!30}84.7  & 70.6 \\
\multicolumn{1}{c}{} &Sketch & 88.6 & 86.5 & 83.6 & \cellcolor{gray!30}84.3 &  &Grass  & 76.7 & 81.3 & 85.9 & \cellcolor{gray!30}61.9 \\
\hline
\multirow{4}{*}{LwF~\cite{li2017lwf}}    &Art              & \cellcolor{gray!30}88.2 & 86.8 & 83.7 & 83.8 &  &Eating  & \cellcolor{gray!30}76.4 & 84.7 & 89.3 & 69.6 \\
                     &Cartoon & 89.4 & \cellcolor{gray!30}86.2 & 85.8  & 84.7 &  &Ground  & 76.6  & \cellcolor{gray!30}82.4 & 84.8 & 67.3   \\
                     &Photo & 86.9 & 86.9  & \cellcolor{gray!30}83.3 & 85.7 &  &Water  & 74.3 & 81.7 & \cellcolor{gray!30}85.5 & 70.7 \\
                     &Sketch & 88.3 & 86.6 & 83.4 & \cellcolor{gray!30}84.3  &  &Grass  & 76.9 & 81.1 & 85.6 & \cellcolor{gray!30}62.4 \\
                     \hline
\multirow{4}{*}{LUCIR~\cite{hou2019lucir}}    &Art            & \cellcolor{gray!30}88.1 & 87.1 & 83.8 & 84.4  &  &Eating  & \cellcolor{gray!30}76.5 & 84.9 & 89.5 & 69.4 \\
                     &Cartoon & 89.6 & \cellcolor{gray!30}86.6 & 85.6  & 85.3 &  &Ground  & 76.8 &\cellcolor{gray!30} 82.1 & 85.5 & 67.4 \\
                     &Photo & 87.0 & 87.1  & \cellcolor{gray!30}83.5 & 85.1 &  &Water  & 73.5  & 82.2 & \cellcolor{gray!30}84.2 & 69.5 \\
                     &Sketch & 88.5 & 86.8 & 83.3 & \cellcolor{gray!30}83.8 &  &Grass  & 77.2 & 81.0 & 85.9 & \cellcolor{gray!30}62.1 \\
                     \hline
\multirow{4}{*}{GeoDL~\cite{simon2021geodl}} &Art                 & \cellcolor{gray!30}87.0 & 86.7 & 83.2 & 85.6 &  &Eating  & \cellcolor{gray!30}72.8 & 86.8 & 86.5  & 67.5 \\
                     &Cartoon & 88.3 &\cellcolor{gray!30} 85.0 & 84.8 & 85.4 &  &Ground  & 76.2 & \cellcolor{gray!30}83.6 & 87.3 & 64.1 \\
                     &Photo & 87.6 & 87.5 & \cellcolor{gray!30}82.0 & 86.5 &  &Water  & 73.0 & 86.4 & \cellcolor{gray!30}85.4 & 68.4 \\
                     &Sketch & 88.4 & 87.4 & 84.0 & \cellcolor{gray!30}84.3 &  &Grass  & 77.6 & 85.9 & 87.2 & \cellcolor{gray!30}60.5  \\
                     \hline
 \multirow{4}{*}{ARM~\cite{zhang2020arm}}         &Art         &\cellcolor{gray!30} 86.2 & 85.6 & 82.6 & 86.5 &  & Eating & \cellcolor{gray!30}65.1 & 83.8 & 80.1 & 66.0 \\
                     &Cartoon & 86.2  & \cellcolor{gray!30}83.5 & 84.2 & 83.4 &  &Ground  & 73.5 & \cellcolor{gray!30}83.3 & 85.4 & 63.7 \\
                     &Photo & 88.1 & 85.6 & \cellcolor{gray!30}80.8  & 84.4 &  & Water  & 72.5 & 87.6 & \cellcolor{gray!30}84.9 & 69.8 \\
                     &Sketch & 85.0 & 87.3 & 81.9 & \cellcolor{gray!30}83.5 &  & Grass & 72.3 & 83.4 & 83.6 & \cellcolor{gray!30}64.9  \\
                     \hline
 \multirow{4}{*}{MixStyle~\cite{zhou2021mixstyle}}     &Art        & \cellcolor{gray!30}86.1  & 85.1 & 83.8 & 81.6  &  & Eating & \cellcolor{gray!30}73.3 & 85.8 & 86.6 & 68.6 \\
                     &Cartoon & 89.4 & \cellcolor{gray!30}84.0 & 84.5 & 84.0 &  & Ground & 75.1 & \cellcolor{gray!30}82.6 & 86.1 & 61.3 \\
                     &Photo & 86.8 & 86.1 & \cellcolor{gray!30}81.0  & 84.1 &  & Water  & 70.5  & 85.6 & \cellcolor{gray!30}81.1 & 68.5 \\
                     &Sketch & 88.6 & 86.6 & 82.8 &\cellcolor{gray!30} 83.4 &  & Grass & 76.9 & 82.1 & 87.3 &\cellcolor{gray!30} 63.6 \\
                     \hline
 \rowcolor{aliceblue}     &Art          & \cellcolor{gray!30}89.9 & 88.9 & 84.9 & 85.8 &  &Eating  & \cellcolor{gray!30}\textbf{77.3} & 85.9 & 90.0  & 69.2 \\
    \rowcolor{aliceblue}                  &Cartoon & 89.6 & \cellcolor{gray!30}86.2  & 85.7 & 84.3 &  &Ground  & 81.6 &\cellcolor{gray!30} 82.1 & 91.2 & 69.3 \\
     \rowcolor{aliceblue}                 &photo & 90.0 & 87.0 & \cellcolor{gray!30}84.4 & 84.5 &  &Water  & 77.8 & 84.8 &\cellcolor{gray!30} 87.0 & 75.0 \\
     \rowcolor{aliceblue}   \multirow{-4}{*}{MSL  (ours) }              &Sketch & 88.6 & 88.3 & 84.4 & \cellcolor{gray!30}85.1 &  &Grass  & 80.5 & 84.1 & 90.2 & \cellcolor{gray!30}62.8 \\
     \hline
 \rowcolor{aliceblue} &Art         & \cellcolor{gray!30}\textbf{91.3} & 89.8 & 86.3 & 88.2 &  &Eating  & \cellcolor{gray!30}{77.2} & 85.4  & 90.3 & 68.6 \\
           \rowcolor{aliceblue}           &Cartoon & 91.4 & \cellcolor{gray!30}\textbf{87.9} & 86.3 & 86.7 &  &Ground  & 76.4 &\cellcolor{gray!30} \textbf{84.1} & 88.0 & 65.5 \\
        \rowcolor{aliceblue}              &Photo & 91.0 & 90.2 & \cellcolor{gray!30}\textbf{85.0}     & 86.5 &  &Water  & 76.7 & 85.6  & \cellcolor{gray!30}\textbf{89.0} & 66.9 \\
       \rowcolor{aliceblue}  \multirow{-4}{*}{MSL + Mov  (ours) }             &Sketch & 89.9 & 89.8 & 86.5 & \cellcolor{gray!30}\textbf{87.2} &  &Grass  & 77.2 & 84.7 & 89.5  & \cellcolor{gray!30}\textbf{64.9}\\
       \hline
\end{tabular}
}\caption{Cross-domain continual learning average accuracy with the 2-tasks protocol, and 5 exemplar size in the memory on PACS and NICO datasets. Within each block that corresponds to a method, the domain shown in the beginning of each row is excluded from training, while each column shows the achieved accuracy when a test set of a domain is used. Therefore, the grey cells show the accuracy on unseen domains in leave-one-domain-out protocol while the off-diagonal cells are the accuracy numbers on the test set of seen domains.}
\label{tab:seendomain_pacs_nico}
\end{table*}


\end{document}